%% file: main.tex
\documentclass{article}

\usepackage{natbib}
\usepackage[final]{neurips_2025}

\usepackage{floatflt}

\usepackage{amssymb}
\usepackage[utf8]{inputenc} 
\usepackage[T1]{fontenc}    
\usepackage{hyperref}       
\usepackage{url}            
\usepackage{booktabs}       
\usepackage{amsfonts}       
\usepackage{nicefrac}       
\usepackage{microtype}      
\usepackage{xcolor}         
\usepackage{wrapfig}

\usepackage{float,wrapfig}
\usepackage{subcaption}
\usepackage{ulem}

\usepackage{multirow}

\usepackage{balance}
\usepackage{xspace}

\definecolor{citypink}{RGB}{227, 108, 194}

\newcommand{\OM}{OmniSVG\xspace}
\newcommand{\OBench}{MMSVGBench}
\newcommand{\ODataset}{MMSVG-2M}

\usepackage{amsmath}
\usepackage[capitalize]{cleveref}
\crefname{section}{Sec.}{Secs.}
\Crefname{section}{Section}{Sections}
\Crefname{table}{Table}{Tables}
\crefname{table}{Tab.}{Tabs.}
\renewcommand{\paragraph}[1]{\vspace{0.1em}\noindent\textbf{#1}}

\usepackage{multirow}
\usepackage{xcolor}
\usepackage{colortbl}
\usepackage[most]{tcolorbox}
\usepackage{adjustbox}
\usepackage{pifont}
\usepackage{makecell}
\usepackage{array}
\usepackage{tikz}

\definecolor{my_red}{RGB}{204, 0, 0}
\title{\OM: A Unified Scalable Vector Graphics Generation Model}

\author {
    \textbf{Yiying Yang}\textsuperscript{\rm 1,2}$^*$
    \quad
    \textbf{Wei Cheng}\textsuperscript{\rm 2}$^*$
    \quad
    \textbf{Sijin Chen}\textsuperscript{\rm 1}
    \quad
    \textbf{Xianfang Zeng}\textsuperscript{\rm 2}
    \quad 
    \textbf{Fukun Yin}\textsuperscript{\rm 1,2}
    \\
    \textbf{Jiaxu Zhang}\textsuperscript{\rm 2}
    \quad
    \textbf{Liao Wang}\textsuperscript{\rm 2}
    \quad
    \textbf{Gang Yu}\textsuperscript{\rm 2 \ddag}
    \quad
    \textbf{Xingjun Ma}\textsuperscript{\rm 1 \ddag}
    \quad
    \textbf{Yu-Gang Jiang}\textsuperscript{\rm 1} \\[0.5em]
    $^{1}$ Fudan University
    \quad
    $^{2}$ StepFun \\[0.5em]
    \textcolor{citypink}{\small
    \raisebox{-0.2\height}{\includegraphics[height=0.5cm]{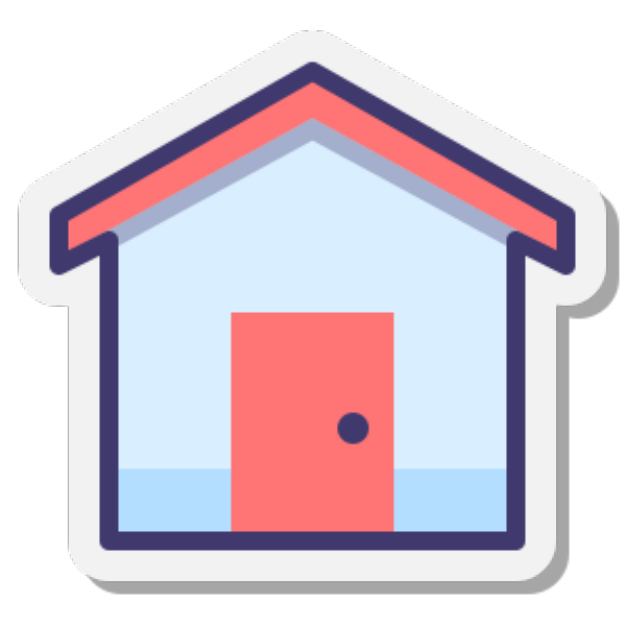}}~{\href{https://omnisvg.github.io/}{\textbf{Project Page}}}
    \quad
    \raisebox{-0.2\height}{\includegraphics[height=0.5cm]{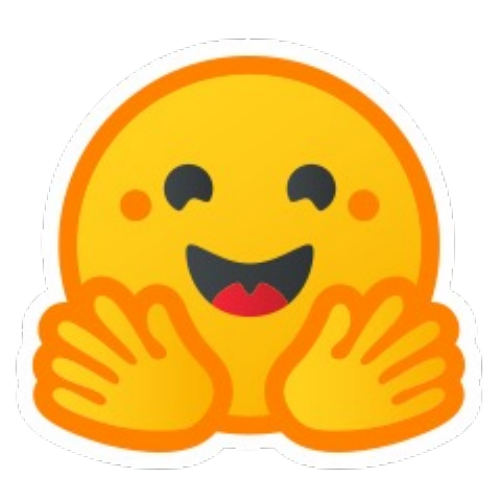}}~{\href{https://huggingface.co/OmniSVG/datasets}{\textbf{\ODataset}}}
    \quad
    \raisebox{-0.2\height}{\includegraphics[height=0.5cm]{files/huggingface_logo.pdf}}~{\href{https://huggingface.co/datasets/OmniSVG/MMSVGBench}{\textbf{\OBench}}}
    \quad
    \raisebox{-0.2\height}{\includegraphics[height=0.5cm]{files/huggingface_logo.pdf}}~{\href{https://huggingface.co/OmniSVG}{\textbf{Models}}}
    \quad
    \raisebox{-0.2\height}{\includegraphics[height=0.5cm]{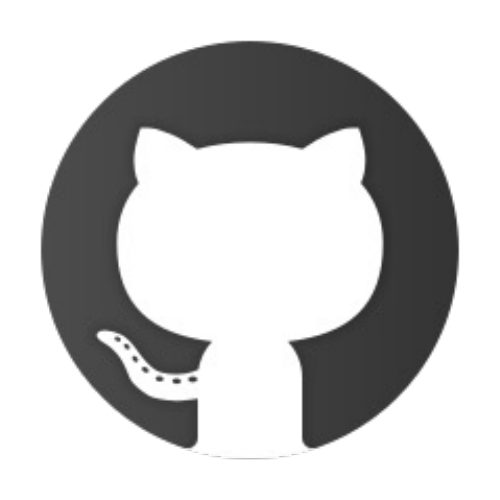}}~{\href{https://github.com/OmniSVG/OmniSVG}{\textbf{Code}}}
    }
}

\begin{document}

\maketitle

{\let\thefootnote\relax\footnotetext{\noindent *~Yiying Yang and Wei Cheng contributed equally to this work.}
{\let\thefootnote\relax\footnotetext{\noindent \ddag~Corresponding Authors.}}

\input{Figures/1.teaser}

\input{Sections/0.abstract}

\clearpage

\section{Introduction}\label{sec:intro}
\input{Sections/1.intro}

\section{Related Works}\label{sec:related}

\input{Sections/2.related}

\section{OmniSVG Dataset}\label{sec:MMSVGBench}
\input{Sections/3.1MMSVG_bench}

\section{OmniSVG}\label{sec:OmniSVG }
\input{Sections/3.2OmniSVG}

\section{Experiments}\label{sec:exp}
\input{Sections/4.exp}

\section{Conclusions}\label{sec:conclusion}

\input{Sections/6.conclusion}

\clearpage
{
    \small
    \balance
    \bibliographystyle{plain}
    \normalem
    \bibliography{main}
}


\newpage

\textbf{\LARGE Appendix}
\appendix

\setcounter{section}{0}
\makeatletter

\input{Sections/7.appendix}

\end{document}

%% file: Figures/1.teaser.tex
\begin{figure*}[h]
    \centering
    \vspace{-20px}
    \includegraphics[width=\linewidth]{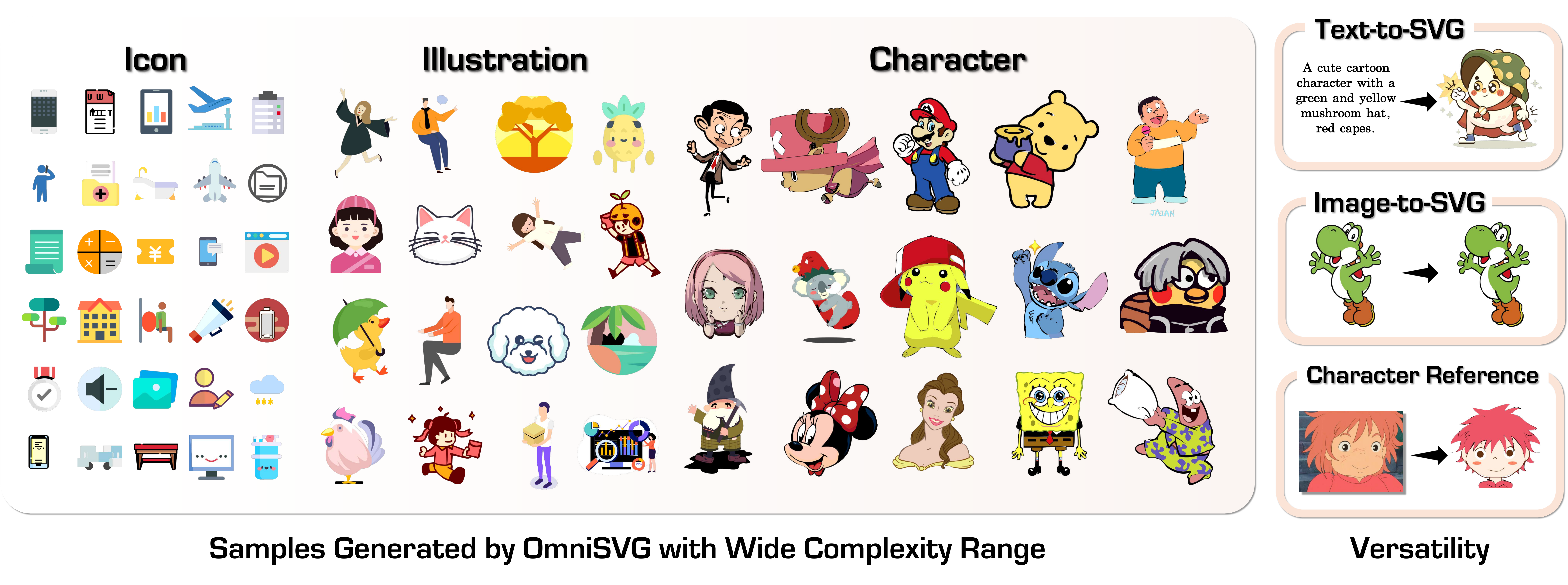}
    \caption{\small \textbf{\OM} is capable of autoregressively generating high-quality Scalable Vector Graphs (SVG) across a wide spectrum of complexity, from simple icons to intricate anime characters. 
    \OM demonstrates remarkable versatility in generating high-quality SVGs adhering to multimodal instructions, covering tasks like Text-to-SVG, Image-to-SVG, and Character-Reference SVG, making it a powerful and flexible solution for diverse creative tasks.}
    \label{fig:teaser}
\end{figure*}

%% file: Sections/0.abstract.tex
\begin{abstract}

Scalable Vector Graphics (SVG) is an important image format widely adopted in graphic design because of their resolution independence and editability. 
The development of autonomous SVG generation workflows is continuously drawing attention from both designers and researchers in the AIGC community.
However, existing methods either produce unstructured outputs at huge computational cost or are limited to generating monochrome icons of over-simplified structures. 
To produce high-quality and complex SVG adhering to multi-modal instructions, we propose \textbf{OmniSVG}, a unified SVG generation framework that inherits knowledge from a pre-trained Vision-Language Model (VLM). 
By parameterizing SVG commands and coordinates into discrete token sequences, the auto-regressive nature enables us to seaminglessly adapt modern VLMs to the direct SVG generation.
To further advance the development of SVG synthesis, we introduce \textbf{MMSVG-2M}, a multimodal dataset with two million richly annotated SVG assets, along with a standardized evaluation protocol for conditional SVG generation tasks. 
Extensive experiments show that \OM outperforms existing methods and demonstrates its potential for integration into professional SVG design workflows.

\end{abstract}

%% file: Sections/1.intro.tex
\noindent Scalable Vector Graphics (SVG) have become a cornerstone of modern digital design because of their resolution independence, compact file size, and inherent editability.  
Widely adopted in professional workflows from UI/UX design to industrial CAD systems, SVG enables precise manipulation of geometric primitives (\textit{e.g.}, Bézier curves, polygons) while maintaining high precision and consistent visual quality across varying resolutions. 
However, creating high-quality SVG content remains challenging for non-experts, requiring mastery of specialized tools or intricate XML syntax.

Existing methods adopt either optimization-based methods or auto-regressive approaches to generate SVG contents. 

The optimization-based methods~\citep{live,vtracer,diffvg} iteratively refine the SVG parameters by minimizing the differences between the input image and the raster image created by differentiable vector graphics rasterizers.
Though these methods are sufficient for reconstructing SVG icons, they suffer from significant computational overhead when scaling up to more intricate samples and produce unstructured outputs with redundant anchor points, harming the editability of the reconstructed SVG samples.
In contrast, auto-regressive methods build transformer models or adapt pre-trained Large Language Models (LLMs) to directly generate XML parameters~\citep{llm4svg} or codes~\citep{chat2svg, starvector} representing SVGs. 
Benefiting from the end-to-end learning pipeline, the auto-regressive method is a more scalable approach~\citep{meshxl} as it can learn directly from a large collection of SVG samples.
However, existing auto-regressive approaches are limited to basic SVG contents~\citep{figr, stack, deepvecfont} because of the limited context length and the scarcity of complex SVG data.

In this paper, we propose \textbf{\OM} that harnesses native VLMs~\cite{qwen2.5-VL} for various end-to-end multimodal SVG generation tasks.
By parameterizing SVG coordinates and commands into discrete tokens, \OM decouples structural logic from low-level geometry, mitigating the ``coordinate hallucination'' problem prevalent in code-based LLMs, and produces vivid and colorful SVG results.
Additionally, the next token prediction training objective enables \OM to complete SVGs with diverse generation results given some partial observations. 
Compared to traditional auto-regressive SVG generation methods, \OM is able to parameterize SVGs exceeding $30k$ tokens, facilitating the generation of detailed and complex SVG contents.
Building upon pre-trained VLMs, our method natively integrates the ability to reason upon visual and textual instructions to synthesize editable, high-fidelity SVGs across diverse domains, from icons to intricate illustrations and anime characters.

To advance the development of SVG synthesis, we introduce \textbf{MMSVG-2M}, a multi-modal SVG synthesis dataset with two million richly annotated assets, encompassing icons, illustrations, and anime designs. 

We also establish a standardized evaluation protocol, \textbf{MMSVG-Bench}, for ``Text-to-SVG'' and ``Image-to-SVG'' generation. 
%
Extensive experiments show that \OM can produce highly detailed and complex SVG contents, surpassing prior art both quantitatively and qualitatively.

To summarize, our key contributions include:

\begin{itemize}
\setlength{\leftmargin}{0pt}

\item We introduce \OM, a family of end-to-end multimodal SVG generators that leverage native VLMs for generating complex and detailed SVGs, from simple icons to intricate anime characters.

\item We present MMSVG-2M, a large-scale dataset comprising two million SVG assets, along with a standardized evaluation protocol for various multi-modal SVG generation tasks, providing a comprehensive resource for future research.

\item Extensive experiments show that \OM surpasses prior SVG generation methods both qualitatively and quantitatively, highlighting its potential for integration into professional SVG design workflows.

\end{itemize}

%% file: Sections/2.related.tex
\noindent \textbf{SVG Generation}.
Early attempts to generating SVGs directly utilize architectures like RNNs~\citep{sketchrnn_david_2018,im2vec, supersvg_hu_2024, ClipGen_Shen_2022, CLIPVG_song_2023}, VAEs~\citep{deepsvg,fonts_svg,strokenuwa_tang_2024, marvel_su_2023, evolution_tian_2022}, and Transformers~\citep{deepsvg, iconshop} to compress SVG commands into latent representations. 
Meanwhile, DeepSVG~\citep{deepsvg} further parameterizes SVGs using a dual transformer architecture but struggles with geometric consistency. 
Recently, the advent of large language models (LLMs)~\citep{llava_Liu_2023,yi_young_2024,qwen2_wang_2024,qwen2.5, meshxl, ll3da, shapegpt, pminr, strokenuwa} unleashes the potential of generating SVGs via XML code synthesis~\citep{llm4svg,chat2svg,starvector}. 
%
%
However, the limited context length of existing LLM-based SVG generation methods~\citep{chat2svg, starvector, llm4svg} poses significant challenges in handling complex SVGs that exceed $10k$ tokens. 
In this paper, we explore the potential of native Vision-Language Models (VLMs) in multi-modal SVG generation. 
By combining pre-trained VLMs with SVG command parameterization, we validate that \OM is able to follow multi-modal instructions and generate vivid and complex SVGs.

\noindent \textbf{Image Vectorization}. Recent advancements in vectorization harness diffusion models paired with differentiable rasterizers, using techniques like score distillation sampling~\citep{dreamfusion,vectorfusion,svgbuilder} and specialized regularizers~\cite{diffvg,live} to convert raster images into SVG paths. 
While these methods achieve remarkable results, they face limitations such as over-smoothing, color over-saturation, and lack of editability, often producing tangled paths that fail to capture hierarchical structures inherent in professional SVG designs. 
In this paper, we present an end-to-end approach that follows multi-modal instructions to generate high-quality SVGs with improved path clarity and editability.

\noindent \textbf{SVG Datasets and Benchmarks}. The lack of suitable datasets for complex SVG structures presents a significant challenge. 
Existing datasets~\citep{figr, stack, deepvecfont} primarily focus on simplified path-based SVGs or monochrome icons, overlooking the intricate layered structures and rich color semantics found in real-world designs. 
For example, FIGR-8-SVG~\citep{figr} focuses on monochromatic icons, while StarVector~\citep{starvector} proposes categorized datasets, including illustrations, icons, emojis, and fonts. 
Therefore, existing datasets only present simple SVG samples that do not exceed $8.2k$ tokens, failing to capture the complexities of layered structures and rich color semantics. 
Benchmark evaluations, such as VGBench~\citep{vgbench}, further highlight gaps in multi-format testing and the absence of comprehensive coverage for illustrative SVGs.
To this end, we introduce \textbf{MMSVG-2M}, a multimodal SVG synthesis dataset comprising two million richly annotated assets, including icons, illustrations, and complex anime designs. 
We also present a standardized evaluation protocol, \textbf{MMSVG-Bench}, to evaluate diverse multi-modal SVG generation tasks with varying complexity.

%% file: Sections/3.1MMSVG_bench.tex
\noindent 
We present \textbf{MMSVG-2M}, a large-scale SVG dataset with two million SVG samples covering website icons, illustrations, graphic designs, anime characters, and etc (\cref{subsec:mmsvgbench_data_curation}). 
To promote the downstream development of SVG generation methods, we also introduce \textbf{MMSVG-Bench}, a standardized evaluation protocol for a series of multi-modal instruction following tasks for conditional SVG generation (\cref{subsec:mmsvgbench_task}).

\subsection{MMSVG-2M} 
\label{subsec:mmsvgbench_data_curation}

\paragraph{Data Source.} 
With increasing visual complexity, MMSVG-2M consists of three subsets, 1) the icon subset MMSVG-Icon collected from Iconfont, 2) the illustration subset MMSVG-Illustration sourced from IconSount, and 3) the complex anime character subset MMSVG-Character both curated from Freepik and created by our data creation pipeline as shown in \cref{fig:pipeline}.
All these websites are online platforms where users can publish and share SVGs, encompassing a broad variety of categories. 
Specifically, our collection of MMSVG-2M contains $1.1$ million icons, $0.5$ million illustrations, and 0.4 million anime characters as shown in \cref{tab:dataset_stastics}.

\paragraph{Data Curation.}
We extract SVG samples with a comprehensive deduplication process based on filenames, SVG code, and metadata. 
We first fit the collected SVGs within a viewbox of $200 \times 200$.
Then, we employ an off-the-shelf VLM, specifically BLIP-2~\citep{blip2}, to generate captions for the SVGs.
Please find more samples from the MMSVG-2M dataset in \cref{fig:MMSVG-2M dataset}, and instruction templates in \cref{appendix: SVG-Image-Text Pairs Construction}.

\paragraph{SVG Simplification} is an essential procedure in SVG data cleansing, since the over-complicated XML grammars in the crawled SVG data will lead to ambiguities while representing basic shapes. To standardize training and evaluation, we simplify all SVG commands with atomic commands as shown in \cref{tab:base_commands}. Inspired by FIGR-8-SVG~\cite{figr} and IconShop~\citep{iconshop}, we remove all attributes and simplify each SVG with five basic commands, including ``Move To'' (M), ``Line To'' (L), ``Cubic Bézier'' (C), ``Elliptical Arc'' (A), ``ClosePath'' (Z). The introduction of atomic commands further removes the ambiguities, as complex XML grammars can be approximated with the combination of several atomic commands.
To efficiently produce a unified and less complex data structure, we utilize picosvg~\footnote{\url{https://github.com/googlefonts/picosvg}} to remove grammars like ``group'' and ``transform'', and simplify the complex commands to atomic path commands. It is worth noting that atomic path commands are sufficient to represent complex SVGs shown in \cref{fig:teaser}.

\input{Tables/3.1SVG_command}

\subsection{MMSVG-Bench}  
\label{subsec:mmsvgbench_task}
    To compensate for the vacancy of standardized and open evaluation for SVG generation, we introduce \textbf{MMSVG-Bench}, a comprehensive benchmark for multi-modal SVG generation.
    We require the corresponding benchmark to be a sufficient verification whether a model is practically useful in real-world scenarios, and avoid the excessive similarity between the benchmark input data and training data as in traditional train/test splits.
    Therefore, we opt to generate the benchmark inputs with GPT-4o.
    Specifically, for Text-to-SVG task, we synthesize 150 textual prompts for each SVG subset (\textit{i.e.} Icon and Illustration). For Image-to-SVG task, we synthesize extra 150 textual descriptions, and prompt GPT-4o to generate vector-style images with transparent backgrounds based on the above texts as the ground truth visual samples.
    We focus on both the visual quality and semantics of the generation results.

\paragraph{Text-to-SVG} requires a model to generate SVGs from text instructions. 
We measure the visual quality with Frechet Inception Distance (FID)~\citep{fid}, aesthetic appeal with Aesthetic score~\citep{aesthetic}, text-SVG alignment with CLIP score~\citep{clip_score}, and Human Preference Scores (HPS)~\citep{hps}.

\paragraph{Image-to-SVG} evaluates a model's ability to convert images into SVGs. 
To quantify the distance between the input and output SVG, we calculate the cosine similarity of DinoV2 features (DinoScore)~\citep{dinov2}, Structural Similarity Index (SSIM)~\citep{ssim}, Learned Perceptual Image Patch Similarity (LPIPS)~\citep{lpips}, and Mean Squared Error (MSE).

\paragraph{Character-Reference SVG Generation} evaluates a model's ability to generate novel SVGs while keeping the profile of the characters depicted in the input image. 
Different from image-to-SVG, the model does not reconstruct, but generates a specific character SVG for the input image (see \cref{fig:cref}).
We evaluate the alignment between input character images and generated SVGs by prompting GPT-4o~\citep{GPT-4o} to generate a score ranging from 1 to 10, the higher the better. ~\cite{gloeckle2024better,katharopoulos2020transformers,guo2025log}

%% file: Tables/3.1SVG_command.tex
\begin{table}[t]
\centering
\caption{\small \textbf{SVG Draw Commands}. Draw commands used in this work along with their arguments and a visualization are listed. The start-position $(x_1$, $y_1)$ is implicitly defined as the end-position of the preceding command.}
\vspace{4mm}
\label{tab:base_commands}
\setlength{\tabcolsep}{0.25em}
\renewcommand{\arraystretch}{1.1}
   \resizebox{0.92\linewidth}{!}{
    \begin{tabular}{>{\centering\arraybackslash} m{1.55cm}>{\centering\arraybackslash} m{2.3cm}>{\centering\arraybackslash} m{5.6cm}>{\centering\arraybackslash} m{3.9cm}}
    \toprule
    \textbf{Command} & \textbf{Arguments} & \textbf{Description} & \textbf{Visualization} \\
    \midrule
        \texttt{<SOP>} &
        $\varnothing$ &
        \begin{tabular}{l}
            'Start-of-Path' token.
        \end{tabular} &
        \\
    \midrule
        \begin{tabular}{c}
            \texttt{M} \\
            (MoveTo)
        \end{tabular} &
        $x_2$, $y_2$ &
        \begin{tabular}{l}
            Move the cursor to the end-point $(x_2, y_2)$ \\
            without drawing anything.
        \end{tabular} &
        \includegraphics[width=1.5cm]{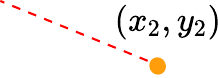} \\
    \midrule
        \begin{tabular}{c}
            \texttt{L} \\
            (LineTo)
        \end{tabular} &
        $x_2$, $y_2$ & 
        Draw a line to the point $(x_2, y_2)$. &
        \includegraphics[width=2cm]{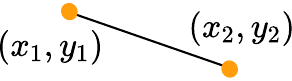} \\
    \midrule
        \begin{tabular}{c}
            \texttt{C} \\
            (Cubic \\
            Bézier)
        \end{tabular}
        \vspace{0pt} &
        \begin{tabular}{l}
             $q_{x1}$, $q_{y1}$ \\ $q_{x2}$, $q_{y2}$ \\
             $x_2$, $y_2$
        \end{tabular}
        \vspace{0pt} &
        \begin{tabular}{l}
            Draw a cubic Bézier curve with control \\
            points $(q_{x1}, q_{y1})$, $(q_{x2}, q_{y2})$ and end-point \\
            $(x_2, y_2)$.
        \end{tabular}
        \vspace{0pt} &
        \includegraphics[width=3cm]{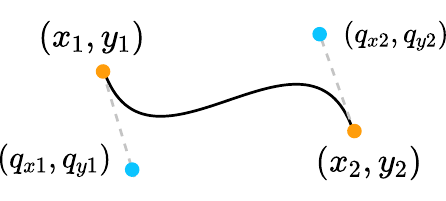} \\
        \midrule
        \begin{tabular}{c}
            \texttt{A} \\
            (Elliptical \\
            Arc)
        \end{tabular}
        \vspace{0pt} &
        \begin{tabular}{l}
             $r_{x}$, $r_{y}$\\ $\varphi$, $f_{A}$, $f_{S}$\\ 
             $x_{2}$, $y_{2}$
        \end{tabular}
        \vspace{0pt} &
        \begin{tabular}{l}
            Draw an elliptical arc with radii $r_x$ and $r_y$\\ (semi-major and semi-minor axes), \\ rotated by angle $\varphi$ to the x-axis, \\ and end-point ($x_2$, $y_2$).
            $(x_2, y_2)$.
        \end{tabular} &
        \includegraphics[width=2.5cm]{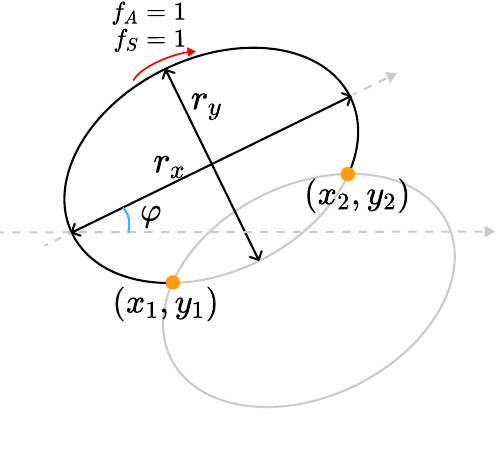} \\
    \midrule
        \texttt{Z} (ClosePath) & 
        $\varnothing$ &
        \begin{tabular}{l}
            Close the path by moving the cursor back \\
            to the path's starting position $(x_0, y_0)$.
        \end{tabular} 
        &
        \includegraphics*[trim=0 6 0 6,width=1.8cm]{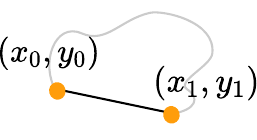} \\
    \midrule
        \texttt{F} (Fill) & 
        $fill$ &
        \begin{tabular}{l}
            Draw the fill attribute of the path. 
        \end{tabular} 
        &
        $\varnothing$ \\

    \midrule
        \texttt{<EOS>} &
        $\varnothing$ &
        \begin{tabular}{l}
            'End-of-SVG' token.
        \end{tabular} &
         \\
    \bottomrule
    \end{tabular}}\vspace{-3mm}
\end{table}

%% file: Sections/3.2OmniSVG.tex
\noindent To support end-to-end training for multi-modal SVG generation, \OM parameterizes a series of atomic SVG path commands into a sequence before feeding into a pre-trained VLM with multi-modal instructions.

\noindent \textbf{SVG Tokenizer.} As illustrated in Sec.~\ref{sec:MMSVGBench}, our MMSVG-2M dataset simplifies an SVG by removing all attributes and using five basic path commands (see \cref{tab:base_commands}).
After the simplification, an SVG script $G$ is represented as the combination of $M$ paths, $G=\{P_{i}\}_{i=1}^{M}$. 
Here, $P_i$ is the $i$-th path containing $N_i$ commands, $P_{i}=\{C_{i}^{j}\}_{j=1}^{N_{i}}$, where $C_{i}^{j}$ is the $j$-th command in the $i$-th path. 
Each command is represented as $C_{i}^{j}=(U_{i}^{j}, V_{i}^{j})$, containing both the command type identifier $U_{i}^{j} \in\{\mathrm{M}, \mathrm{L}, \mathrm{C}, \mathrm{A}, \mathrm{Z}\}$ and the corresponding location argument $V_{i}^{j}$.

To generate colored SVG contents, we assign special tokens for hex values to control the ``Fill'' ($\mathrm{F}$) attribute, distinguishing it from the original SVG commands and coordinates. 
To this end, we are able to use a total six types of commands $U_{i}^{j} \in\{\mathrm{M}, \mathrm{L}, \mathrm{C}, \mathrm{A}, \mathrm{Z}, \mathrm{F}\}$ to parameterize a colored SVG parameterization.

Specifically, our SVG tokenizer transforms SVG scripts $X_s$ into an ordered SVG token sequence within the same representation space as the pre-trained VLM. 
Following IconShop~\citep{iconshop}, we flatten the layered structure of the SVG script by concatenating different paths into a single command sequence, where each path begins with the drawing commands followed by point coordinates. 
Therefore, each SVG sequence could be represented as a flattened sequence.
As the generation identifier, we apply special tokens like \texttt{<SOP>} and \texttt{<EOS>} to the two ends of a SVG sequence, identifying the begining and ending of a SVG sequence.
We assign special tokens for each command type, \textit{i.e.} $\{\mathrm{M}, \mathrm{L}, \mathrm{C}, \mathrm{A}, \mathrm{Z}, \mathrm{F}\}$.
To shorten the length of the SVG sequence, we further merge the 2D point coordinates into one token with a mapping function: $ <x,y> \rightarrow x\times w + y$, where $w$ is the width of the image.
The SVG sequence are then lifted into the same embedding space as the pre-trained VLM with a learnable embedding layer.

\input{Figures/2.pipeline}

\noindent \textbf{Model Architecture.}  OmniSVG adopts Qwen2.5-VL~\cite{qwen2.5-VL}, an open-sourced VLM that excels at understanding intricate vision-text inputs, as its backbone (\cref{fig:pipeline}) to produce precise and compact SVG outputs. 
OmniSVG is trained to predict the SVG suffix tokens ($x_s$) conditioned on the mutli-modal instruction prefix tokens ($x_c$) with the standard next-token prediction objective.

\begin{equation}
\vspace{-4mm}
\theta^{*} = \arg \max_{\theta} \prod_{i = 1}^{L} P\left(x_{s, i} \mid x_{s,<i}, x_{c}\right)
\end{equation}

%% file: Figures/2.pipeline.tex
\begin{figure*}
    \centering
    \includegraphics[width=\linewidth]{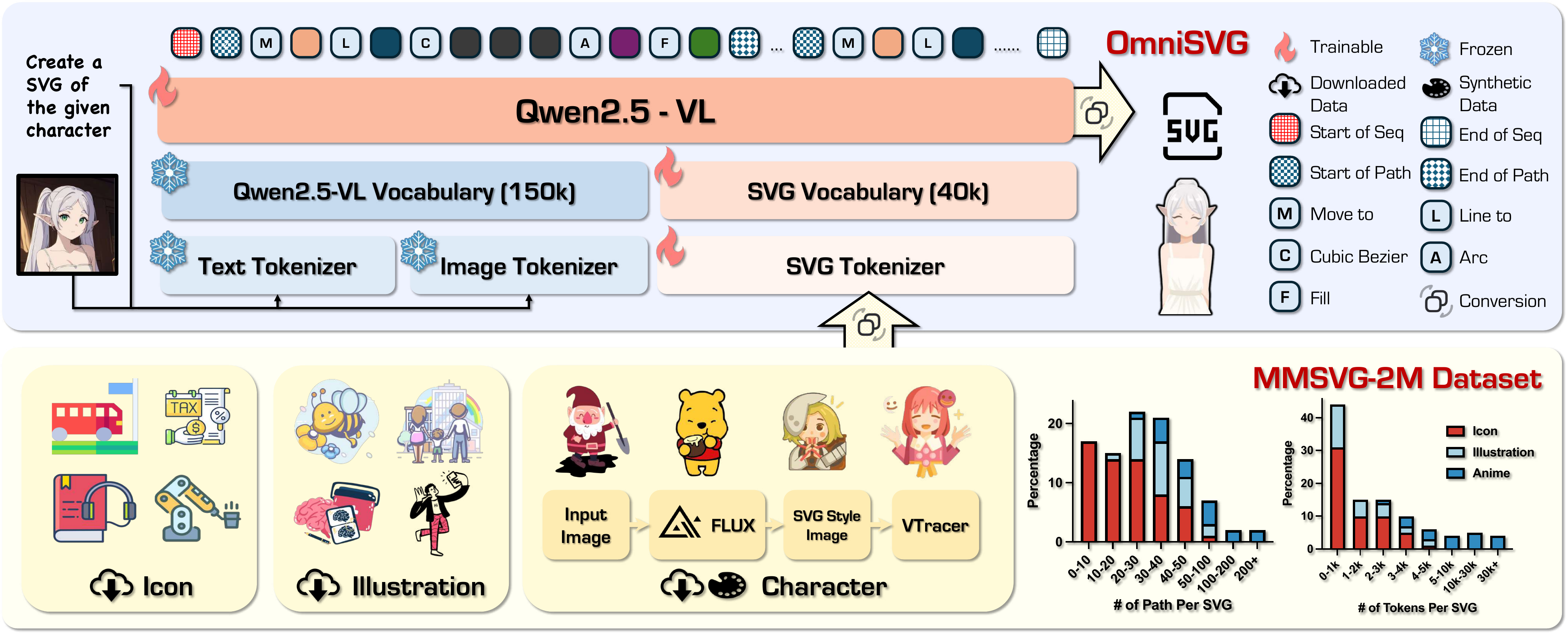}
    \caption{\small \textbf{Overview of \OM}. \OM~is built on a pre-trained vision-language model Qwen2.5-VL and incorporates an SVG tokenizer. The model tokenizes both text and image inputs as prefix tokens, while the SVG tokenizer encodes vector graphics commands into a unified representation space. }
    \label{fig:pipeline}
    \vspace{-5mm}
\end{figure*}

%% file: Sections/4.exp.tex
\input{Tables/1.quantative_sota}

\noindent To validate the effectiveness of our method, we first introduce the baselines (Sec. \ref{subsec: baseline}). 
Then, we make quantitative comparisons with prior arts (\cref{subsec: qualitative comparison,subsec: quantitative comparison}) and conduct ablations (Sec. \ref{subsec: ablation}) to study the effectiveness of our design.

\subsection{Baselines}
\label{subsec: baseline}


\noindent For the text-to-SVG task, we compare our method with language-based (LLM-based) methods, including VectorFusion~\citep{vectorfusion}, SVGDreamer~\citep{svgdreamer}, Chat2SVG~\citep{chat2svg} and IconShop~\citep{iconshop}. For image-to-SVG task, we compare our method with baseline methods across image vectorization and Multimodal Large Language Modeling approaches, including LIVE~\citep{live}, DiffVG~\citep{diffvg}, StarVector~\citep{starvector}, Vtracer~\citep{vtracer} and GPT-4o~\citep{GPT-4o} using the official implementations with the hyperparameters proposed by the authors, and apply their pre- and post-processing code as required.

\subsection{Quantitative Comparisons} 
\label{subsec: quantitative comparison}

%
    We compare our OmniSVG with other baseline methods on the ``text-to-SVG'' and ``image-to-SVG'' tasks in our MMSVG-Bench. 
    In addition to the metrics mentioned in \cref{sec:MMSVGBench}, we also report the average token length (\# tokens) of a generated SVG sample utilizing the Qwen2.5-VL~\citep{qwen2.5-VL} tokenizer.

As shown in Tab.~\ref{tab:quantitative_sota}, OmniSVG demonstrates strong performance compared to state-of-the-art baselines in text-to-SVG generation, achieving superior FID scores and competitive CLIP score, aesthetic quality, and HPS.
%
    For image-to-SVG, OmniSVG also achieves competitive results with traditional vectorization methods, \textit{i.e.} LIVE~\cite{live}, DiffVG~\cite{diffvg}, and VTracer~\cite{vtracer}, but with a much shorter sequence length.
    When comparing to auto-regressive methods, \textit{i.e.} GPT-4o~\cite{GPT-4o} and StarVector~\cite{starvector}, OmniSVG also achieves a superior performance across all metrics.
    The above results indicate that OmniSVG effectively balances the generation cost and the visual quality when generating SVGs according to multi-modal conditions.
%

\subsection{Qualitative Evaluations} 
\label{subsec: qualitative comparison}

\paragraph{Text-to-SVG task}. We compare our method with baseline approaches using seven distinct text prompts for the text-to-SVG task, as shown in \cref{fig:Qualitative image-to-SVG Evaluations}. 
Optimization-based methods like SVGDreamer~\citep{svgdreamer} and VectorFusion~\citep{vectorfusion} require significant computation time due to their iterative optimization processes, which, while effective for refining SVG details, are computationally expensive. 
Auto-regressive methods, such as IconShop~\citep{iconshop} and Chat2SVG~\citep{chat2svg}, generate SVGs more quickly by leveraging pre-trained models but have notable limitations. 
IconShop produces monochrome SVGs, restricting its applicability, while Chat2SVG, though flexible, generates less detailed and semantically consistent SVGs in its first stage. 
Our OmniSVG consistently outperforms all baselines across various text prompts in generating high-fidelity SVGs with rich color, geometric accuracy, and the ability to handle complex visual cues.

\input{Figures/4.quality_results}

\input{Figures/ip_gallery}
\paragraph{Image-to-SVG Task.} We compare our method with classical image vectorization approaches, including DiffVG~\citep{diffvg}, LIVE~\citep{live}, and VLM-based methods GPT-4o~\citep{GPT-4o}, StarVector~\citep{starvector} and Vtracer~\citep{vtracer} 
As shown in \cref{fig:Qualitative image-to-SVG Evaluations}, our method outperforms these baselines in both quality and efficiency. 
Optimization-based methods like DiffVG and LIVE perform well on simple icons but struggle with complex images, often generating visual artifacts. 
The GPT-4o model, while capable of generating SVGs for complex images, is limited to icon-level outputs and cannot handle detailed illustrations. 
StarVector excels at simple icons but fails to produce accurate SVGs for more intricate images, highlighting its limited generalization capability. 
Vtracer is an image processing algorithm designed to convert raster images into SVGs. 
In contrast, OmniSVG efficiently converts a wide range of images, from icons to complex illustrations and character images, into high-quality, editable SVGs. 
This superior performance in handling diverse visual cues distinguishes OmniSVG from traditional vectorization methods. 
Additional visual results can be found in \cref{fig:Gallery}. We provide more detailed discussions with existing methods, particularly the recent works LLM4SVG~\cite{llm4svg} and StarVector~\cite{starvector}, in the \cref{appendix: more comparisons of the baselines}.

\paragraph{Character-Reference SVG generation task.} As shown in \cref{fig:cref}, by training on MMSVG-Character with natural character image and SVG pair data, OmniSVG is capable of generating character SVGs through image references.

\subsection{Ablation studies}
\label{subsec: ablation}

\input{Tables/2.3ablation}

\noindent \textbf{Effectiveness of SVG Parameterization.} We present a comprehensive comparison among different SVG parameterization strategy with the traditional non-parameterized methods for SVG representation in large language models. 
We ablates on the parameterization on both coordinate and color attributes of the SVG.

The results, shown in \cref{tab: SVG parameterization} and \cref{fig:svg_param} demonstrate that parameterizing both coordinate and color attributes yields a better generation results under all metrics with the shortest token length. 
It further validates that the efficient token representation allows our method to generate complex SVGs with fewer computational resources. 
Additionally, qualitative results show that our method outperforms others, particularly as SVG complexity increases. The non-parameterization method fails to generate SVGs for complex images. 
These findings underscore the effectiveness of our full parameterization strategy in balancing performance and resource efficiency for SVG generation tasks.

\input{Tables/2.2ablation}

\noindent \textbf{Ablation studies on model size.} To analyze whether training a larger model benefits SVG generation, we evaluate OmniSVG base models with different sizes on the MMSVG-2M dataset in \cref{tab: ablation of the model size}. 
We evaluate OmniSVG with base models of varying sizes on the MMSVG-2M dataset in Tab.~\ref{tab: ablation of the model size} by progressively scaling up the model size. 
The results show that as the model size grows, we can generate SVG samples with a better quality.

\input{Tables/2.1ablation}

\noindent \textbf{Ablation Studies on the VLM Architecture.} To evaluate the effectiveness of the VLM architecture, we conducted an ablation study replacing it with alternative LLM-based architectures incorporating image encoders such as CLIP ViT-B/32~\citep{clip-vit/32}, VQGAN~\citep{vqgan}, and Qwen2.5-VL~\citep{qwen2.5-VL}.

\input{Figures/5.ablation_parm}

The results in \cref{tab: ablation of the VLM architecture} show that Qwen2.5-VL consistently outperformed all alternatives under all evaluation metrics.

\input{Tables/user_study}

\textbf{User Study.} We extract one-tenth of the samples from the evaluation dataset and conducted a user study with 15 participants to evaluate user preferences, vividness, and the alignment between text-to-SVG and image-to-SVG. Participants are asked to assess SVGs generated by different models based on 150 text descriptions and 150 image prompts, comparing the results generated using our method and baseline models. The results in \cref{tab:userstudy} show that OmniSVG is widely preferred, with higher scores for vividness and superior semantic alignment with the input conditions.

%% file: Tables/1.quantative_sota.tex
\begin{table*}
\small
\caption{\small \textbf{Quantitative Evaluations.} Quantitative results between OmniSVG and current state-of-the-art text-to-SVG and image-to-SVG baseline methods. The bold numbers and underlined numbers represents the best and second best performance repectively. Our OmniSVG model demonstrates superior performance compared SOTA SVG generation baselines.}
\label{tab:quantitative_sota}
\centering
\setlength{\extrarowheight}{0pt}
\addtolength{\extrarowheight}{\aboverulesep}
\addtolength{\extrarowheight}{\belowrulesep}
\setlength{\aboverulesep}{0pt}
\setlength{\belowrulesep}{0pt}
\resizebox{\linewidth}{!}{
\begin{tabular}{cccccccccccc} 
\toprule
\multirow{2}{*}{\textbf{Evaluation Dataset}}                                                                     & \multirow{2}{*}{\textbf{Methods}} & \multirow{2}{*}{\textbf{\# Tokens}} & \multicolumn{4}{c}{\textbf{Text-to-SVG}}                                 & \multicolumn{4}{c}{\textbf{Image-to-SVG}}                              \\ 
\cmidrule(l){4-7} \cmidrule(l){8-11}
                                                                                                                 &                                   &                                &                                        FID$\downarrow$ & CLIP$\uparrow$  & Aesthetic$\uparrow$ & HPS$\uparrow$  & DINO$\uparrow$ & SSIM$\uparrow$ & LPIPS$\downarrow$ & MSE$\downarrow$  \\ 
\midrule

\rowcolor[rgb]{0.937,0.941,0.945} {\cellcolor[rgb]{1,1,1}}                                           & Vectorfusion~\cite{vectorfusion}                                                                           & 66.2k                                    & 250.77           &  0.240          & \textbf{4.76}               & 0.237          & –              & –              & –                 & –                \\
\rowcolor[rgb]{0.937,0.941,0.945} {\cellcolor[rgb]{1,1,1}}                                           & SVGDreamer~\cite{svgdreamer}                                                                          & 132.0k                                   & 308.94           & 0.207          & 4.26                & 0.221          & –              & –              & –                 & –                \\
\rowcolor[rgb]{0.937,0.941,0.945} {\cellcolor[rgb]{1,1,1}}                                           & Chat2SVG~\cite{chat2svg}                                                                              & 0.6k                                   & 190.87           & \textbf{0.299}          & 4.41                & \textbf{0.247}         & –              & –              & –                 & –                \\
\rowcolor[rgb]{0.937,0.941,0.945} {\cellcolor[rgb]{1,1,1}}                                           & IconShop~\cite{iconshop}                                                                             & 2.0k                                   & 213.28           & \underline{0.288}          & 4.55                & \underline{0.244}          & –              & –              & –                 & –                \\
\rowcolor[rgb]{0.937,0.941,0.945} {\cellcolor[rgb]{1,1,1}}                                           & LIVE~\cite{live}                                                                                 & 52.5k                                  & –               & –               & –                   & –              & 0.932          & 0.943 & 0.106    & 0.011   \\
\rowcolor[rgb]{0.937,0.941,0.945} {\cellcolor[rgb]{1,1,1}}                                           & DiffVG~\cite{diffvg}                                                                               & 322.0k                                  & –               & –               & –                   & –              & \underline{0.940}          & \underline{0.954}          & 0.066             & \textbf{0.002}            \\
\rowcolor[rgb]{0.937,0.941,0.945} {\cellcolor[rgb]{1,1,1}}                                           & GPT-4o~\cite{GPT-4o}                                                                        & 0.3k                                   & –               & –               & –                   & –              & 0.860          & 0.792          & 0.403             & 0.124            \\
\rowcolor[rgb]{0.937,0.941,0.945} {\cellcolor[rgb]{1,1,1}}                                           & StarVector(8B)~\cite{starvector}                                                                   & 2.0k                                   & –               & –               & –                   & –              & 0.895          & 0.881          & 0.231             & 0.059            \\
\rowcolor[rgb]{0.937,0.941,0.945} {\cellcolor[rgb]{1,1,1}}                                           & Vtracer                                                                   & 52.4k                                   & –               & –               & –                   & –              & \textbf{0.993}          & \textbf{0.966}          & \textbf{0.039}             & \underline{0.002}            \\
\rowcolor[rgb]{1,0.961,0.961} {\cellcolor[rgb]{1,1,1}}                                               & OmniSVG(4B)                                                                & 3.8k                                   & \underline{137.40}   & 0.275  & \underline{4.62}        & \underline{0.244}  & \underline{0.993}  & 0.950          & \underline{0.050}             & 0.006            \\
\rowcolor[rgb]{1,0.961,0.961} \multirow{-11}{*}{{\cellcolor[rgb]{1,1,1}}\textbf{MMSVG-Icon}}         & OmniSVG-L(8B)                                                       & 5.7k                                   & \textbf{130.56}  & 0.276 & 4.60       & 0.242 & 0.922 & 0.893  & 0.235     & 0.040    \\ 
\midrule
\rowcolor[rgb]{0.945,0.965,0.925} {\cellcolor[rgb]{1,1,1}}                                          & Vectorfusion~\cite{vectorfusion}                                                                     & 66.1k                                    & 253.94           &   0.185          & \textbf{4.94}                & 0.226          & –              & –              & –                 & –                \\
\rowcolor[rgb]{0.945,0.965,0.925} {\cellcolor[rgb]{1,1,1}}                                          & SVGDreamer~\cite{svgdreamer}                                                                        & 132.0k                                   & 419.70           & 0.201          & 4.37                & 0.221          & –              & –              & –                 & –                \\
\rowcolor[rgb]{0.945,0.965,0.925} {\cellcolor[rgb]{1,1,1}}                                          & Chat2SVG~\cite{chat2svg}                                                                          & 1.0k                                   & 210.03           & \textbf{0.283}          & 4.45                & \textbf{0.250}          & –              & –              & –                 & –                \\
\rowcolor[rgb]{0.945,0.965,0.925} {\cellcolor[rgb]{1,1,1}}                                          & IconShop~\cite{iconshop}                                                                        & 2.6k                                   & \textbf{107.93}           & \underline{0.233}          & 4.46                & 0.224          & –              & –              & –                 & –                \\
\rowcolor[rgb]{0.945,0.965,0.925} {\cellcolor[rgb]{1,1,1}}                                          & LIVE~\cite{live}                                                                         & 52.2k                                  & –               & –               & –                   & –              & 0.935          & 0.950 & 0.111     & 0.008   \\
\rowcolor[rgb]{0.945,0.965,0.925} {\cellcolor[rgb]{1,1,1}}                                          & DiffVG~\cite{diffvg}                                                                        & 322.0k                                  & –               & –               & –                   & –              & \underline{0.945}          & \underline{0.955}          & \underline{0.065}             & \textbf{0.001}            \\
\rowcolor[rgb]{0.945,0.965,0.925} {\cellcolor[rgb]{1,1,1}}                                          & GPT-4o~\cite{GPT-4o}                                                                    & 0.4k                                   & –               & –               & –                   & –              & 0.875          & 0.854          & 0.373             & 0.077            \\
\rowcolor[rgb]{0.945,0.965,0.925} {\cellcolor[rgb]{1,1,1}}                                          & StarVector(8B)~\cite{starvector}                                                            & 2.6k                                   & –               & –               & –                   & –              & 0.877          & 0.900          & 0.238             & 0.046            \\
\rowcolor[rgb]{0.945,0.965,0.925} {\cellcolor[rgb]{1,1,1}}                                          & Vtracer                                                            & 57.6k                                   & –               & –               & –                   & –              & \textbf{0.994}          & \textbf{0.966}          & \textbf{0.035}             & \underline{0.002}            \\
\rowcolor[rgb]{1,0.961,0.961} {\cellcolor[rgb]{1,1,1}}                                              & OmniSVG(4B)                                                         & 5.8k                                   & 154.37   & 0.226  & \underline{4.56}        & \underline{0.232}  & 0.899  & 0.906          & 0.237             & 0.034            \\
\rowcolor[rgb]{1,0.961,0.961} \multirow{-11}{*}{{\cellcolor[rgb]{1,1,1}}\textbf{MMSVG-Illustration}} & OmniSVG-L(8B)                                           & 6.9k                                  & \underline{138.42}  & 0.231 & 4.51       & \underline{0.232} & 0.905 & 0.907  & 0.231    & 0.031    \\

\bottomrule
\end{tabular}
}

\end{table*}

%% file: Figures/4.quality_results.tex
\begin{figure*}
    \centering
\includegraphics[width=\linewidth]{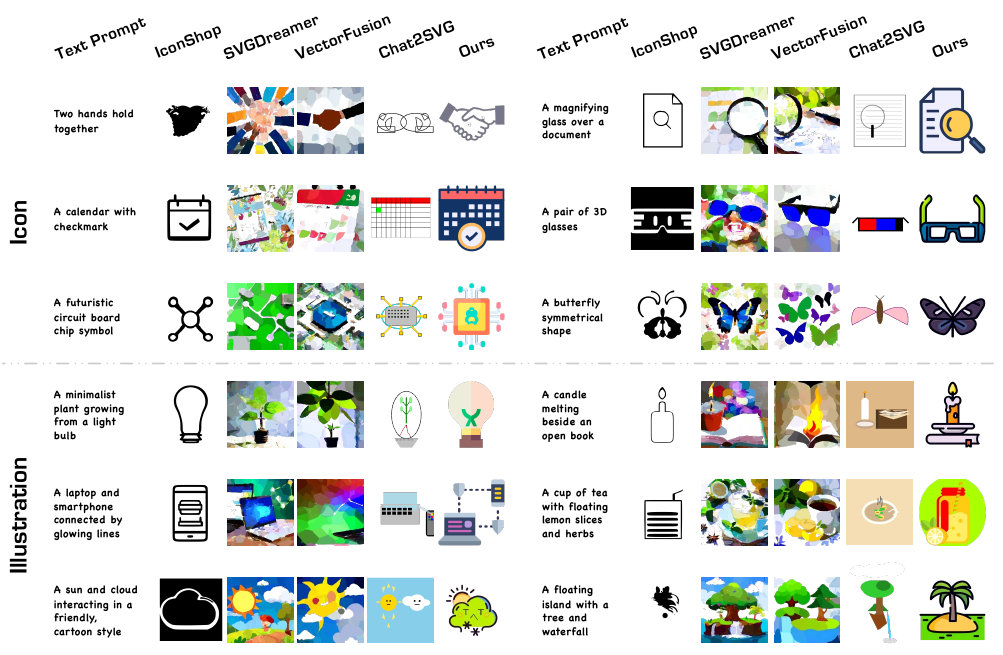}
    \caption{\small \textbf{Qualitative Comparison with SOTA Methods on Text-to-SVG Task}. We compare the propose method with SOTA Text-to-SVG methods on our evaluation benchmarks, namely Icon and Illustration. 
}
    \label{fig:Qualitative text-to-SVG Evaluations}
\end{figure*}

\begin{figure*}
    \centering
    \includegraphics[width=\linewidth]{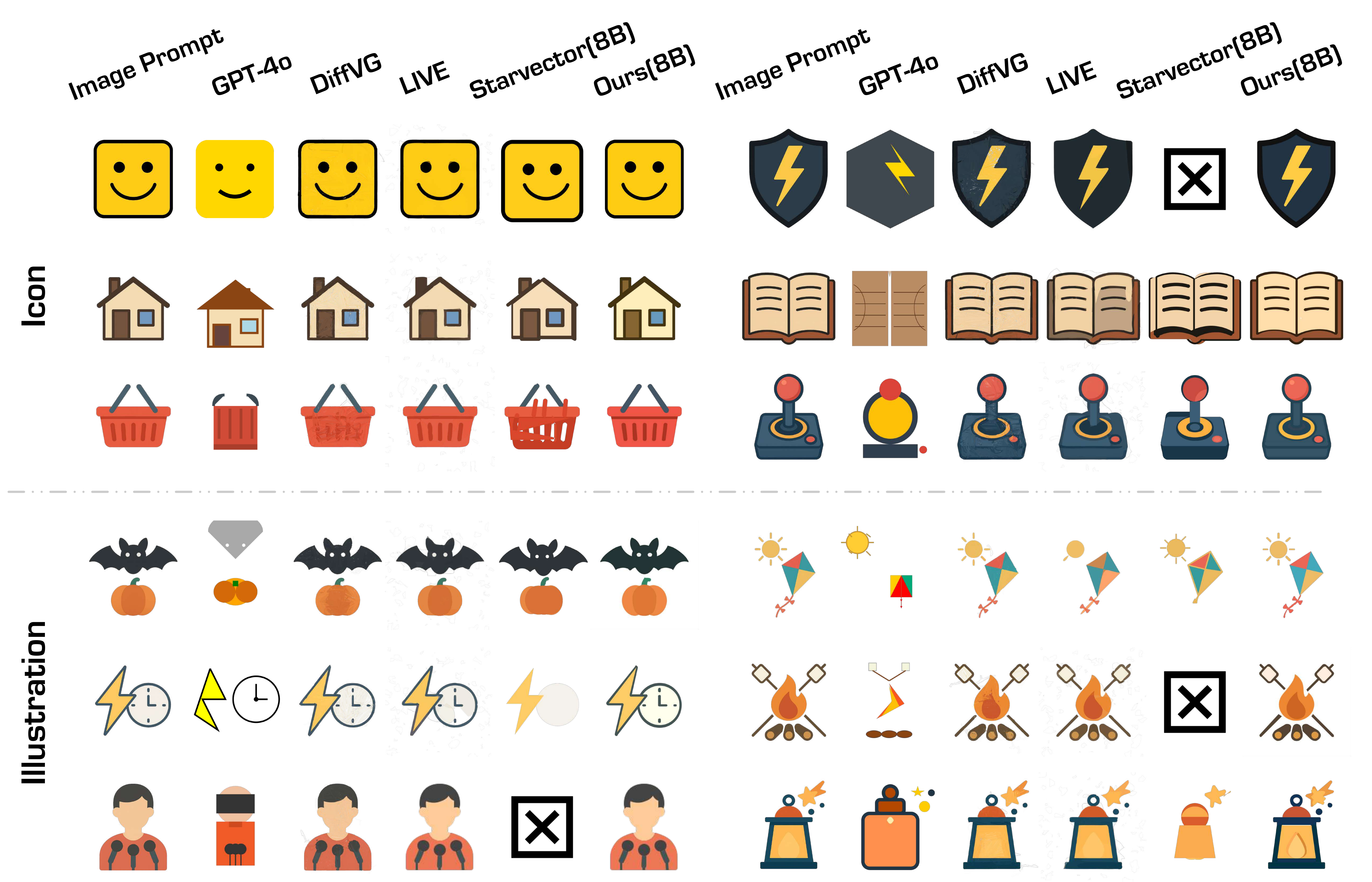}
    \caption{\small \textbf{Qualitative Comparison with SOTA Methods on Image-to-SVG Task}. We compare the propose method with SOTA Image-to-SVG methods on our evaluation benchmarks. 
    }
    \label{fig:Qualitative image-to-SVG Evaluations}
\end{figure*}

%% file: Figures/ip_gallery.tex
\begin{wrapfigure}{r}{0.5\textwidth}
    \centering
    \vspace{-4mm}
    \includegraphics[width=0.85\linewidth]{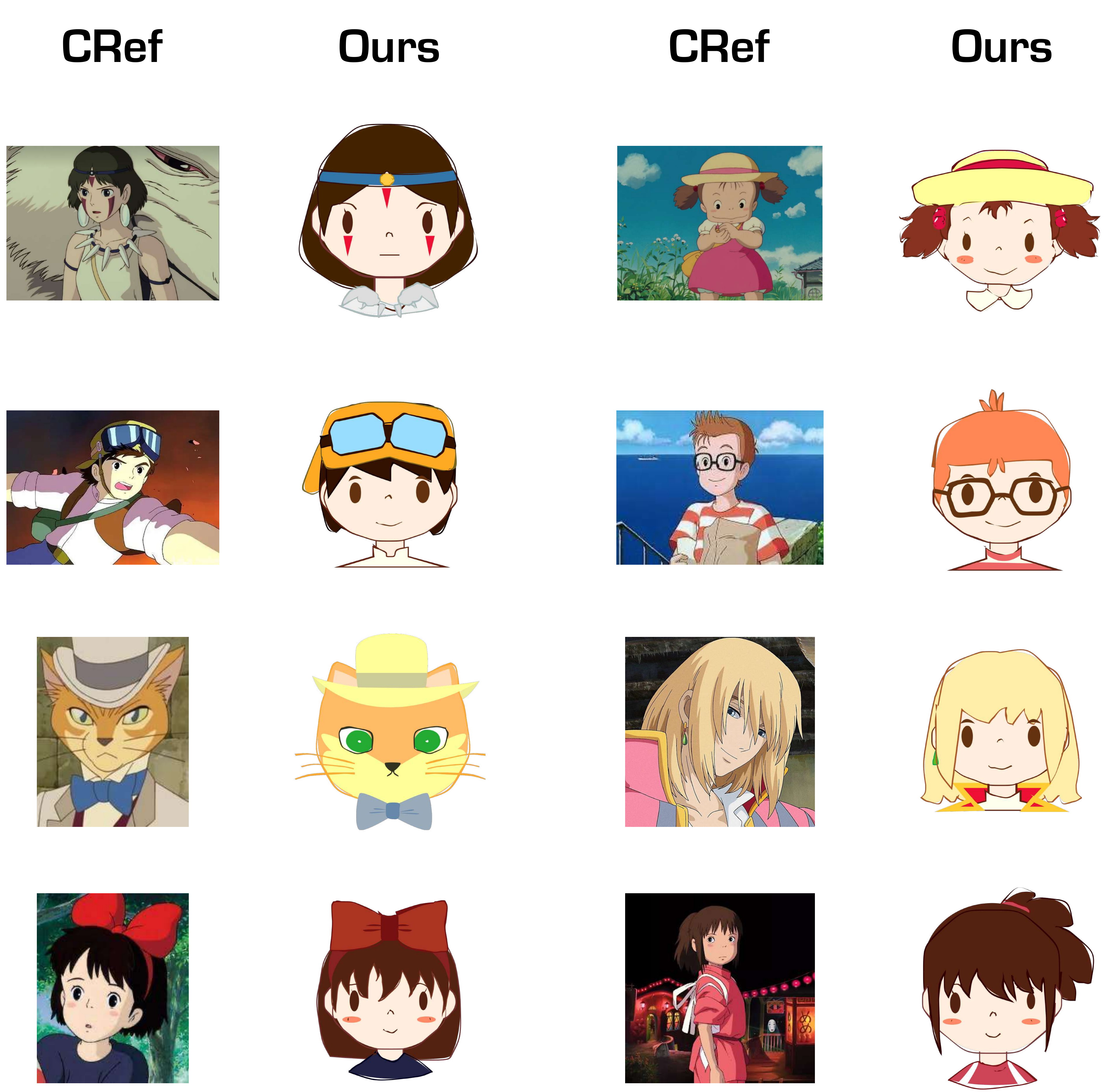}
    \vspace{-2mm}
    \caption{\small \textbf{Generated SVG with Character-Reference (CRef) by OmniSVG.} 
    }
    \vspace{-2mm}
    \label{fig:cref}
\end{wrapfigure}

%% file: Tables/2.3ablation.tex
\begin{table}[b]
 \huge
\caption{\small \textbf{Quantitative Study on SVG Parameterization.}
   Ablation studies on color parametrization (abbreviated as color param.) and coordinate paramterization (abbreviated as coord param.) are conducted.
   }
    \label{tab: SVG parameterization}
    \vspace{3mm}
  \centering
  \resizebox{0.93\linewidth}{!}{
    \scalebox{1}{\begin{tabular}{lccccccccc}
    \toprule
     \multirow{2}{*}{\textbf{Methods}} & \multicolumn{4}{c}{\textbf{Text-to-SVG}} &   \multicolumn{4}{c}{\textbf{Image-to-SVG}} &   \multirow{2}{*}{\textbf{\# Tokens}} \\
      & FID$\downarrow$  & CLIP$\uparrow$ & Aesthetic$\uparrow$ & HPS$\uparrow$  & DINO$\uparrow$ & SSIM$\uparrow$  & LPIPS$\downarrow$ & MSE $\downarrow$ \\
     \midrule

     w/o param.   &  218.76   & 0.185  & 3.43  & 0.138 & 0.741 & 0.718 & 0.315 & 0.182 & 18.5k \\
     w/o coordinate  param.    &  193.42   & 0.216  & 3.90  & 0.169 & 0.826 & 0.809 & 0.248 & 0.119  & 10.2k  \\
      w/o color  param.  & \underline{167.28}   & \underline{0.269}  & \underline{4.31}  & \underline{0.211} & \underline{0.895} & \underline{0.879} & \underline{0.179} & \underline{0.053} & 6.3k  \\
    \textbf{OmniSVG(4B)}    & \textbf{145.89}   & \textbf{0.308}  & \textbf{4.59}  & \textbf{0.238} & \textbf{0.946} & \textbf{0.928} & \textbf{0.138} & \textbf{0.020} & 4.8k  \\
    \bottomrule
    \end{tabular}%
    }
}

\end{table}%

%% file: Tables/2.2ablation.tex
\begin{table}[t]
  \centering
\caption{
   \small \textbf{Ablation of the Model Size.} As the model size grows, the generated samples are of higher quality.}
    \label{tab: ablation of the model size}
    \vspace{2mm}
  \resizebox{0.96\linewidth}{!}{
    \scalebox{1}{\begin{tabular}{lcccccccccc}
    \toprule
     \multirow{2}{*}{\textbf{Methods}} &  \multirow{2}{*}{\textbf{Input}} & \multirow{2}{*}{\textbf{Size}} & \multicolumn{4}{c}{\textbf{Text-to-SVG}} &   \multicolumn{4}{c}{\textbf{Image-to-SVG}} \\
     & & &  FID↓  & CLIP↑ & Aesthetic↑ & HPS↑  & DINO↑ & SSIM↑ & LPIS↓ & MSE↓ \\
     \cmidrule(r){1-1}  \cmidrule(r){2-2} \cmidrule(r){3-3} \cmidrule(r){4-7} \cmidrule(r){8-11}
   FLAN-T5-Base\citep{t5}   &  Text & 223M & 198.48   & 0.158   & 3.38 &  0.085  & -- &   -- & -- & -- \\
    FLAN-T5-Large\citep{t5}   &  Text & 770M & 175.24  & 0.208   & 3.92  & 0.142  & -- &   -- & -- & -- \\
    FLAN-T5-xl\citep{t5}   & Text & 3B & 160.28   & 0.258  & 4.31 & 0.192  & -- &   -- & -- & -- \\
    blip2-flan-t5-xl\citep{blip2}  & Text/Image & 3.94B & 152.11   &  0.235   & 4.48 & 0.215 & 0.898  & 0.891  & 0.255 & 0.041   \\
   \textbf{OmniSVG(4B)}  & Text/Image & 3.7B & \underline{145.89}   & \textbf{0.308}  & \textbf{4.59}  & \textbf{0.238} & \textbf{0.946} & \textbf{0.928} & \textbf{0.138} & \textbf{0.020}  \\
    \bottomrule
    \end{tabular}%
    }
    }
\vspace{-4mm}
\end{table}%

%% file: Tables/2.1ablation.tex
\begin{table}[hbtp]
  \centering
    \caption{
    \small \textbf{Ablation on VLM architecture.} 
   }
\label{tab: ablation of the VLM architecture}
\vspace{2mm}
  \resizebox{0.96\linewidth}{!}{
    \scalebox{1}{\begin{tabular}{cccccccccc}
    \toprule
    \multirow{2}{*}{\textbf{Vision Model}} &  \multirow{2}{*}{\textbf{Language Model}}  & \multicolumn{4}{c}{\textbf{Text-to-SVG}} &   \multicolumn{4}{c}{\textbf{Image-to-SVG}} \\
    & & FID$\downarrow$ & CLIP$\uparrow$ & Aesthetic$\uparrow$ & HPS$\uparrow$ & DINO$\uparrow$ & SSIM$\uparrow$ &  LPIPS$\downarrow$ & MSE$\downarrow$ \\
    
    \cmidrule{1-2} \cmidrule{3-6} \cmidrule{7-10}
    CLIP     & Qwen2.5   & 185.31   & 0.249  & 4.52 & 0.215 & 0.867 & 0.856 & 0.267 & 0.058 \\
    VQGAN     & Qwen2.5   &  198.74  & 0.234   & 4.49 & 0.203 & 0.839 & 0.828 & 0.295 & 0.071 \\ 
    \multicolumn{2}{c}{\textbf{Qwen2.5-VL-3B-Instruct}}  & \underline{145.89}   & \textbf{0.308}  & \textbf{4.59}  & \textbf{0.238} & \textbf{0.946} & \textbf{0.928} & \textbf{0.138} & \textbf{0.020} \\
    \multicolumn{2}{c}{\textbf{Qwen2.5-VL-7B-Instruct}} & \textbf{134.45}   & \underline{0.254}   & \underline{4.56}  & \underline{0.237} & \underline{0.914} &  \underline{0.900} & \underline{0.233} & \underline{0.036}  \\
    \bottomrule
    \end{tabular}
    }
  }
\end{table}

%% file: Figures/5.ablation_parm.tex
\begin{wrapfigure}{r}{0.5\textwidth}
  \centering
  \vspace{-4mm}  
  \includegraphics[width=0.48\textwidth]{{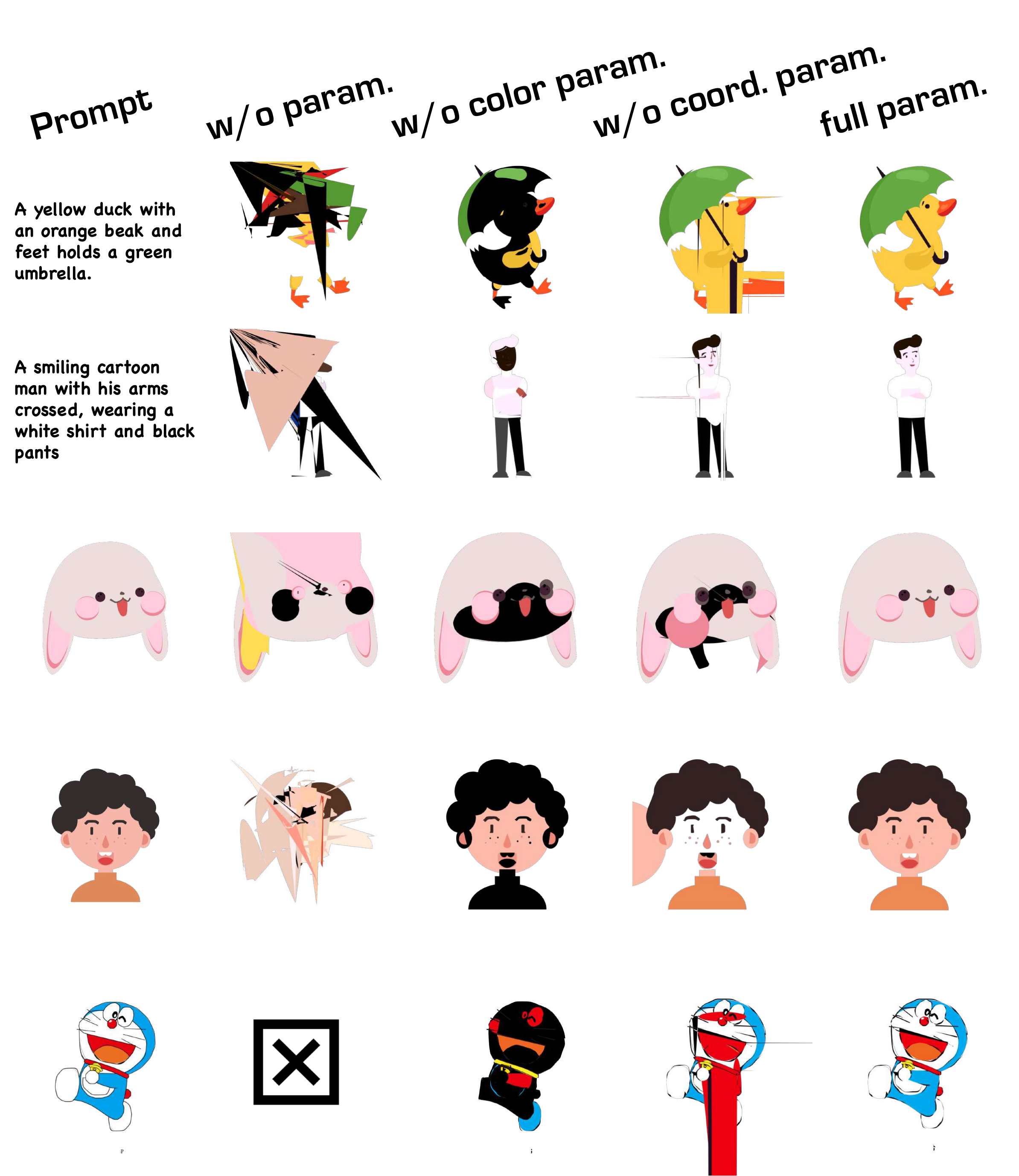}}
    \caption{\small \textbf{Qualitative Study on Parametrization.} }
  \label{fig:svg_param}
  \vspace{-10mm}  
\end{wrapfigure}

%% file: Tables/user_study.tex
\begin{wrapfigure}{r}{\linewidth}
\centering
\vspace{-6mm}
\caption{\small \textbf{User Study of OmniSVG and baselines.}}
    \label{tab:userstudy}
  \resizebox{0.95\linewidth}{!}{
    \scalebox{1}{\begin{tabular}{lccc}
        \toprule
        \textbf{Method}  & \textbf{Preference} ↑ & \textbf{Vividity}↑ & \textbf{Alignment}↑ \\
        \midrule
        Vectorfusion~\cite{vectorfusion} & 35 & 58 & 76 \\
        SVGDreamer~\cite{svgdreamer} & 41 & 65 & 79   \\        
        Chat2SVG~\cite{chat2svg} & 55 & 61 & 86   \\        
        IconShop~\cite{iconshop} & 79 & 57 & 75   \\       
        GPT-4o~\cite{GPT-4o} & 38 & 54 & 80   \\   
        StarVector(8B)~\cite{starvector} & 37 & 81 & 68   \\        
        DiffVG~\cite{diffvg} & 88 & 76 & 96   \\ 
        LIVE~\cite{live} & 86 & 70 & 95  \\
        \textbf{OmniSVG} & \textbf{96} & \textbf{88} & \textbf{98}   \\
        
        \bottomrule
         \vspace{-6mm}

    \end{tabular}
}}
\end{wrapfigure}

%% file: Sections/6.conclusion.tex
\textbf{Conclusions.} We introduce OmniSVG, a unified framework for multimodal SVG generation that leverages pre-trained Vision-Language Models (VLMs). 
By parameterizing SVG commands and coordinates as discrete tokens, OmniSVG efficiently decouples structural logic from geometry, addressing issues like "coordinate hallucination" while maintaining design expressiveness. 
Our method outperforms existing approaches in both quality and efficiency, offering high-quality, editable SVG across various design domains. Additionally, we proposed MMSVG-2M, a large-scale multimodal dataset with two million annotated SVG assets and a standardized evaluation protocol. 
Extensive experiments show that OmniSVG surpasses prior SVG generation methods in various conditional generation tasks, highlighting its potential for integration into professional SVG design workflows.

\textbf{Limitations and Future Work.} 
During inference, OmniSVG generates tens of thousands of tokens for complex samples, which inevitably leads to a considerable generation time. OmniSVG is only bounded by vector style image prompt and fails on natural images.
As for future work, recent endeavors on multi-token prediction~\cite{gloeckle2024better,cai2023medusa} and KV-cache compression~\cite{zhou2024dynamickv,cai2024pyramidkv} provide a promising way to save the generation cost.
Additionally, the auto-regressive nature of OmniSVG also unlocks future opportunities for in-context learning~\cite{zhang2023makes,zhou2024visual,sun2024generative}, chain-of-thought reasoning~\cite{wei2022chain,guo2025deepseek}, and multi-turn interleaved generation~\cite{huang2024dialoggen,liu2024mmdu}, thereby providing a more precise user control.

\section*{Acknowledgements} 
This work is in part supported by National Key R\&D Program of China (Grant No. 2022ZD0160103), National Natural Science Foundation of China (Grant No. 62276067), and National Natural Science Foundation of China (Grant No. 62472104).

The computations in this research were performed using the CFFF platform of Fudan University.

%% file: Sections/7.appendix.tex
\section{Additional Details of MMSVG-2M dataset} 
\label{appendix: Additional Details of our MMSVG-2M dataset}

\subsection{Samples of MMSVG-2M Dataset}
We visualize samples of our MMSVG-2M dataset in \cref{fig:MMSVG-2M dataset}. 
In our MMSVG-2M dataset, 55\% of the SVG samples belongs to the MMSVG-Icon, 25\% belongs to the MMSVG-Illustration, and the rest 20\% belongs to the MMSVG-Character. 
Among the SVG samples within the MMSVG-Character category, half of them comes from Freepik, while another half is generated by our data creation pipeline.
We also collect image-SVG pairs for the character-reference SVG generation tasks during the generation process.

\input{Tables/dataset_statistic}

\subsection{SVG-Image-Text Pairs Construction}
\label{appendix: SVG-Image-Text Pairs Construction}
Our \textbf{\textit{MMSVG-2M}} dataset comprises two million SVG samples with the corresponding rasterized images. 
We generate captions on the rasterized images with BLIP-2~\cite{blip2}, thereby providing textual descriptions that enable us to fine-tune our model to follow these instructions. 
We use CairoSVG~\cite{cairosvg} for rasterization and remove samples that produced completely white images.

\noindent \textbf{Annotation.} We employ an off-the-shelf VLM, specifically BLIP-2~\cite{blip2}, to generate SVG captions with the prompt below. To reduce hallucinations, we drop the samples with CLIP scores less than 30. We also visualize the distribution annotated keywords of MMSVG-2M dataset in Fig.~\ref{fig:wordcloud} with word cloud format. And the instruction template for annotation is shown in \cref{tab:instruction_templates}.

\input{Tables/appendix_template}

\noindent \textbf{Instruction templates.} MMSVGBench provides three tasks, including text-to-SVG task, image-to-SVG task and character-reference SVG generation task. Each task needs different instruction templates. For the text and image conditioning SVG generation, we provide the input text or image with VLM architecture. For character-reference SVG generation, we provide the natural charecter reference image and the original image with the VLM architecture. The list of instruction templates for different tasks are shown in \cref{tab:instruction_templates}.

\subsection{Character-SVG Pairs Construction}
As illustrated in the Fig.~\ref{tab:dataset_stastics}, part of our proposed MMSVG-2M-Character subset is constructed using a generative pipeline. As shown in the pipeline diagram in Fig.~\ref{fig:pipeline}, we employ a FLUX~\citep{flux2024}-based generative model enhanced with a vector-style LoRA to enable the generation of SVG-style data. For image-based conditioning, we adopt FLUX-Redux~\citep{redux}, which injects image features via a SigLIP encoder and projects them into image embeddings. These embeddings are then concatenated with the text tokens as conditioning inputs for FLUX~\citep{flux2024}. However, in practice, the original Redux~\citep{redux} conditioning proves to be overly strong. To address this, we adopt a community-implemented variant of Redux that downsamples the image embeddings in 2D space. As observed in our experiments shown in Fig.~\ref{fig:redux}, a downsampling factor between $2\times$ and $3\times$ yields the most reasonable SVG-style character references. Finally, we employ VTracer~\citep{vtracer} to perform near-instant vectorization of the generated images. To construct the MMSVG-2M-Character subset, we first filter $103k$ character instances from the Danbooru~\citep{Danbooru2023} dataset and apply the aforementioned pipeline with motion and expression keywords like previous works~\cite{cheng2023dna,cheng2022generalizable,pan2023renderme,yu2023monohuman}. We compare the raw FLUX~\citep{flux2024} outputs and their vectorized counterparts, retaining only those samples with PSNR and SSIM scores above a certain threshold as valid data.

\input{Figures/S1.dataset_visualization}

\input{Figures/S1.dataset_redux}

\section{Additional Details}
\label{Additional Details}
\subsection{Scaling Up} To study the effectiveness of scaling up multimodal SVG generation, we scale up OmniSVG from 4B to 8B parameters. We present training perplexity in Fig.~\ref{fig:scale_up}, where both models are trained from scratch on 250 billion tokens. We show that, as the size of the model grows, the model achieves a lower validation perplexity, indicating a higher probability of producing the validation data.

\input{Figures/S1.dataset_wordcloud}

\subsection{Implementation Details} We train our models in bfloat16 with the ZeRO-2 strategy~\cite{zero} for memory-efficient training.  We also adopt the AdamW~\cite{adamw} optimizer with a learning rate decaying from $3 \times 10^{-4}$ to $3 \times 10^{-6}$ and a weight decay of $0.1$ to train our model.  In practice, we load the pre-trained weights from the Qwen2.5-VL~\cite{qwen2.5-VL} model and initialize the SVG embeddings from scratch. Without further specification, we generate SVGs with the top-k and top-p sampling strategy with $k = 50$ and $p = 0.95$ for diversity. 

\input{Figures/3.scale_up}

\section{Additional Results}
\label{Addition Results}
As list in full comparisons in \cref{tab:quantitative_sota}, including all the baselines mentioned in \cref{sec:exp}. For the text-to-SVG task, we compare our method with language-based (LLM-based) methods, including VectorFusion~\citep{vectorfusion}, SVGDreamer~\citep{svgdreamer}, Chat2SVG~\citep{chat2svg} and IconShop~\citep{iconshop}. For image-to-SVG task, we compare our method with baseline methods across image vectorization and Multimodal Large Language Modeling approaches, including LIVE~\citep{live}, DiffVG~\citep{diffvg}, StarVector~\citep{starvector} and GPT-4o~\citep{GPT-4o} using the official implementations with the hyperparameters proposed by the authors, and apply their pre- and post-processing code as required. Specifically, for the text-to-SVG task, the optimization-based method SVGDreamer excels in enhancing editability by employing a semantic-driven image vectorization process that effectively separates foreground objects from the background, while failing to handle complex scenes. Another optimization-based work, VectorFusion, stands out for generating SVG-exportable vector graphics without relying on large captioned datasets. However, Vectorfusion is also unable to handle complex scenarios and diverse styles. The significant problem with these optimization-based works is that the optimization time is too long. Generating an SVG usually takes more than ten minutes, which is too expensive.  For the LLM-based method, Chat2SVG integrates Large Language Models (LLMs) with image diffusion models to create semantically rich SVG templates. However, Chat2SVG still needs to optimize the output SVG script from LLM, which introduces increased computational complexity and poses challenges during model training. In comparison, IconShop utilizes a transformer-based architecture to autoregressively model SVG path sequences, demonstrating exceptional performance in simplified icon SVGs, which offers effective solutions for text-to-SVG generation. It can only generate black simple Icon SVGs.

For the image-to-SVG task, we compare our method with the image vectorization methods. LIVE allows progressive and efficient generation of SVGs, optimizing closed vector paths under raster image supervision with shape complexity control. However, LIVE  needs to optimize for a long time when generating complex SVGs. DiffVG enables end-to-end differentiability in vector graphics rasterization, improving optimization through anti-aliasing and gradient-based methods while also is computationally expensive due to the complexity of the forward-backward rasterization process. Recently, the Multimodal Large Language Model (MLLM) based method StarVector leverages the visual understanding to apply accurate SVG primitive to the LLM architecture, which also can generate SVGs from both text and image inputs. However, it still fails to generate complex SVGs. Since Starvector~\citep{starvector} has not yet opened up its text-to-SVG model weights, our MMSVGBench does not evaluate Starvector's text-to-SVG capabilities. MMSVG-Bench also evaluates our methods with VLM methods, GPT-4o, to conduct a comprehensive assessment. We compare our method with these baselines on our MMSVG-2M dataset, from simple MMSVG-Icon datset, a bit complex MMSVG-illustration dataset, to the very complex MMSVG-Character dataset.

\input{Figures/gallery}

\section{More details of the baselines}
\label{appendix: more comparisons of the baselines}

\subsection{Text-to-SVG Task}
\noindent \textbf{SVGDreamer}~\cite{svgdreamer} uses a semantic-driven image vectorization (SIVE) process to separate foreground objects and background, improving editability. The SIVE process utilizes attention-based primitive control and an attention-mask loss function to manipulate individual elements effectively. To address issues in existing text-to-SVG generation methods, the proposed Vectorized Particle-based Score Distillation (VPSD) approach models SVGs as distributions of control points and colors, improving shape, color diversity, and convergence speed. 

\noindent \textbf{VectorFusion}~\cite{vectorfusion} leverages a text-conditioned diffusion model trained on pixel representations to generate SVG exportable vector graphics without needing large captioned SVG datasets. By optimizing a differentiable vector graphics rasterizer, it distills semantic knowledge from a pretrained diffusion model and uses Score Distillation Sampling to generate an SVG consistent with a caption. Experiments show that VectorFusion improves both quality and fidelity, offering a variety of styles such as pixel art and sketches. 

\noindent \textbf{Chat2SVG}~\cite{chat2svg} proposes a hybrid framework that combines the strengths of Large Language Models (LLMs) and image diffusion models for text-to-SVG generation. The approach first uses an LLM to create semantically meaningful SVG templates from basic geometric primitives. A dual-stage optimization pipeline, guided by image diffusion models, refines paths in latent space and adjusts point coordinates to enhance geometric complexity. 

\noindent \textbf{IconShop}~\cite{iconshop} uses a transformer-based architecture to encode path commands and learn to model SVG path sequences autoregressively. It has shown excellent results in simplified icon scenarios and provides a good solution to Text-to-SVG generation by extending the FIGR-8-SVG dataset with captions. We have access to their dataset and original splits and have trained our model on that data using a pre-trained checkpoint (trained on OmniVG dataset). We have extracted the results from IconShop and included them here to compare our method. 

\noindent \textbf{LLM4SVG}~\cite{llm4svg} is a framework that leverages Large Language Models (LLMs) to understand and generate Scalable Vector Graphics (SVGs). It employs a structured SVG encoding approach, utilizing learnable semantic tokens to accurately represent SVG components and their properties. This design enables LLMs to produce SVGs that are both semantically aligned with textual descriptions and visually coherent. However, LLM4SVG also has a maximum token length of 2048, limiting its ability to generate highly complex SVGs that require longer sequences. 

\subsection{Image-to-SVG Task}
\noindent \textbf{LIVE} (Layer-wise Image Vectorization)~\cite{live} is a method for progressively generating SVGs that closely fit a given raster image by recursively adding and optimizing closed vector paths. Using a differentiable renderer (based on DiffVG~\cite{diffvg}), LIVE enables direct optimization of paths under raster image supervision while controlling shape complexity by adjusting the number of path segments. It introduces component-wise path initialization, identifying key visual components to ensure efficient topology extraction and minimize redundant shapes. 

\noindent \textbf{DiffVG}~\cite{diffvg} is a landmark in vector graphics research, pioneering deep learning-based methods with the first differentiable vector graphics rasterization pipeline. By leveraging a combination of anti-aliasing techniques and gradient-based optimization, DiffVG ensures differentiability. Unlike methods relying on non-differentiable curve-to-mesh conversions, DiffVG employs a forward-backward rasterization process, where the forward pass generates antialiased images and the backward pass computes gradients with respect to vector graphic parameters. 

\noindent \textbf{StarVector}~\cite{starvector} works directly in the SVG code space, leveraging visual understanding to apply accurate SVG primitives. StarVector employs a transformer-based architecture that integrates an image encoder with a language model, enabling it to process visual inputs and produce precise SVG code. StarVector effectively handles diverse SVG types, including icons, logos, and complex diagrams, demonstrating robust generalization across various vectorization tasks. 
However, with a 16k token context window, StarVector may struggle to process highly complex SVGs that require longer sequences.

\noindent \textbf{Vtracer}~\cite{vtracer} is an image processing algorithm designed to convert raster images into SVGs. The algorithm follows a three-step pipeline, which involves the hierarchical clustering of images for vectorization. Initially, the pixels are transformed into paths, which are subsequently simplified into polygons. In the final step, these polygons are smoothed and approximated using a Bezier curve fitter.

%% file: Tables/dataset_statistic.tex
\begin{table}[htbp]
    \centering
    \vspace{-3mm}
\caption{\small \textbf{Data Statistics for MMSVG-2M.} Our MMSVG-2M consists of 1.1 million SVG icons, 0.5 million SVG illustrations, and 0.4 million SVG anime characters.}
    \label{tab:dataset_stastics}
  \resizebox{0.9\linewidth}{!}{
    \begin{tabular}{lccccc}
        \toprule
        \textbf{Dataset}  & \textbf{Train} & \textbf{Val} & \textbf{Total} & \textbf{Source} & \textbf{Token Length} \\
        \midrule
        \textbf{MMSVG-Icon} & 990k & 110k & 1,100k & Iconfont\footnote{\url{https://www.iconfont.cn/}} & $2.2\text{k} \pm 0.9\text{k}$ \\
        \textbf{MMSVG-Illustration} & 450k & 50k & 500k & IconScout & $8.1\text{k} \pm 3.3\text{k}$ \\
        \textbf{MMSVG-Character} & 350k & 50k & 400k & Freepik \& generated & $28\text{k} \pm 7.3\text{k}$ \\
        \bottomrule
    \end{tabular}
  }
  \vspace{-1mm}
\end{table}

%% file: Tables/appendix_template.tex
\begin{table*}[b]
\centering
\begin{tcolorbox}[title=Instructions for Different Tasks,
enhanced,
skin first=enhanced,
skin middle=enhanced,
skin last=enhanced,
colback={rgb,255:red, 255; green, 250; blue, 250},
colframe={rgb,255:red, 227; green, 108; blue, 194},
fonttitle=\bfseries]{}

- \textbf{Employed BLIP2 for SVG Captioning:} You are a helpful assistant. Your task is to describe this image in a single sentence, including the object, its color, and its overall arrangement. For example: ``Yellow cheers with glasses of alcohol drinks." / ``Heart emojis represent love on Valentine's Day." \\

- \textbf{Text-to-SVG:} You are a helpful SVG Generation assistant, designed to generate SVG. We provide  the text description as input, generate SVG based on the text. \\

- \textbf{Image-to-SVG:} You are a helpful SVG Generation assistant, designed to generate SVG. We provide an image as input, generate SVG for this image.  \\

- \textbf{Character-Reference SVG Generation:} 
You are a helpful SVG Generation assistant, designed to generate SVG. We provide a natural image as input, please generate the simplified character SVG based on the reference input image.

\end{tcolorbox}
\caption{\small \textbf{Instructions for Different Tasks}. Instructions including annotation, text-to-SVG, image-to-SVG and character-reference SVG generation.}\label{tab:instruction_templates}
\end{table*}

%% file: Figures/S1.dataset_visualization.tex
\begin{figure*}
    \centering
    \includegraphics[width=0.95\linewidth]{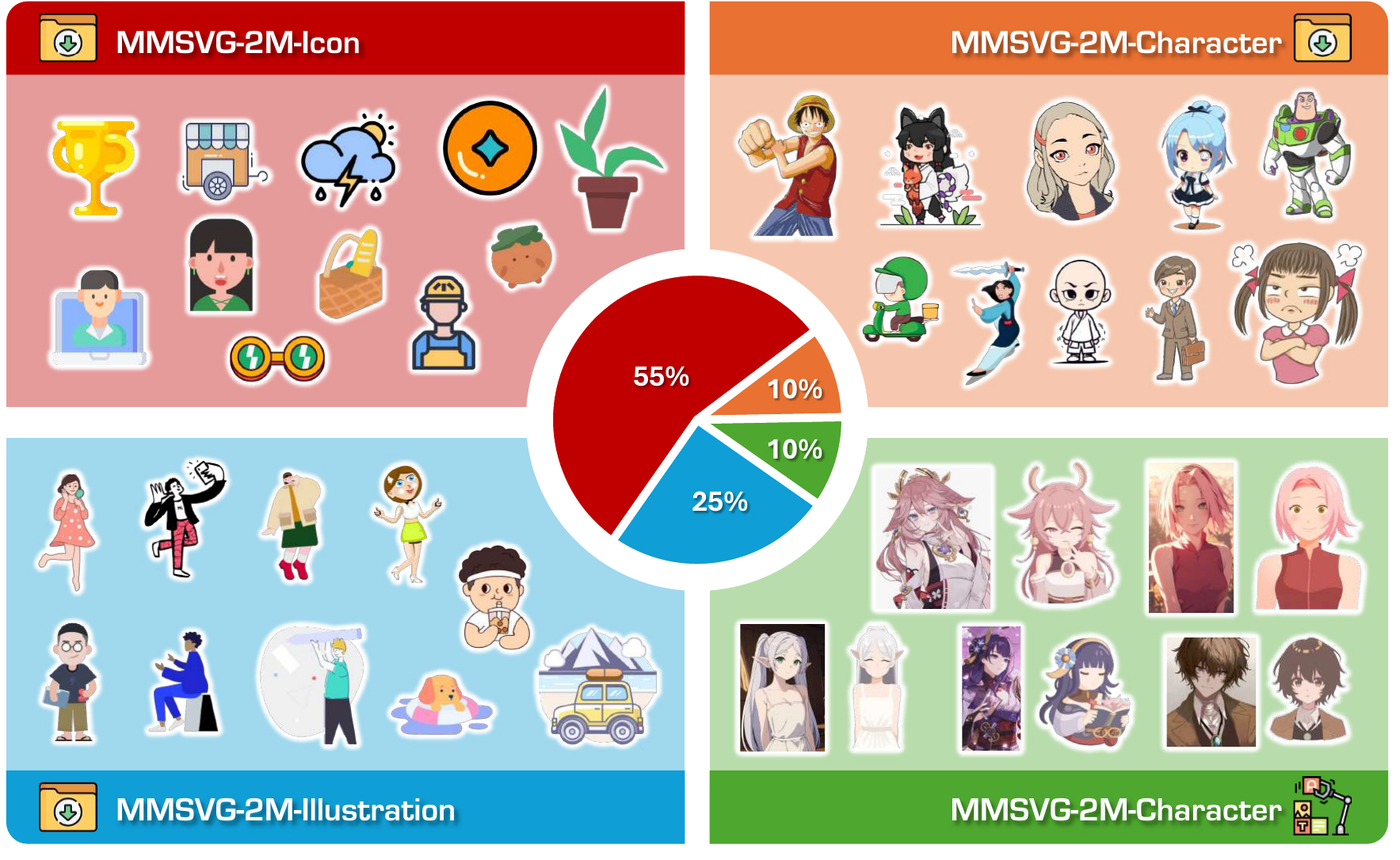}
    \caption{\small \textbf{Samples from MMSVG-2M Dataset}. The proposed MMSVG-2M dataset can be separated into three subset, namely Icon, Illustration and Character. Samples from Icon, Illustration and part of Character subsets are downloaded from Internet. Another part of Character subset is generated by our data creation pipeline, which can provide image and SVG pairs for image prompting task.}
    \label{fig:MMSVG-2M dataset}
    \vspace{-5mm}
\end{figure*}

%% file: Figures/S1.dataset_redux.tex
\begin{figure}[b]
    \centering
    \includegraphics[width=.95\linewidth]{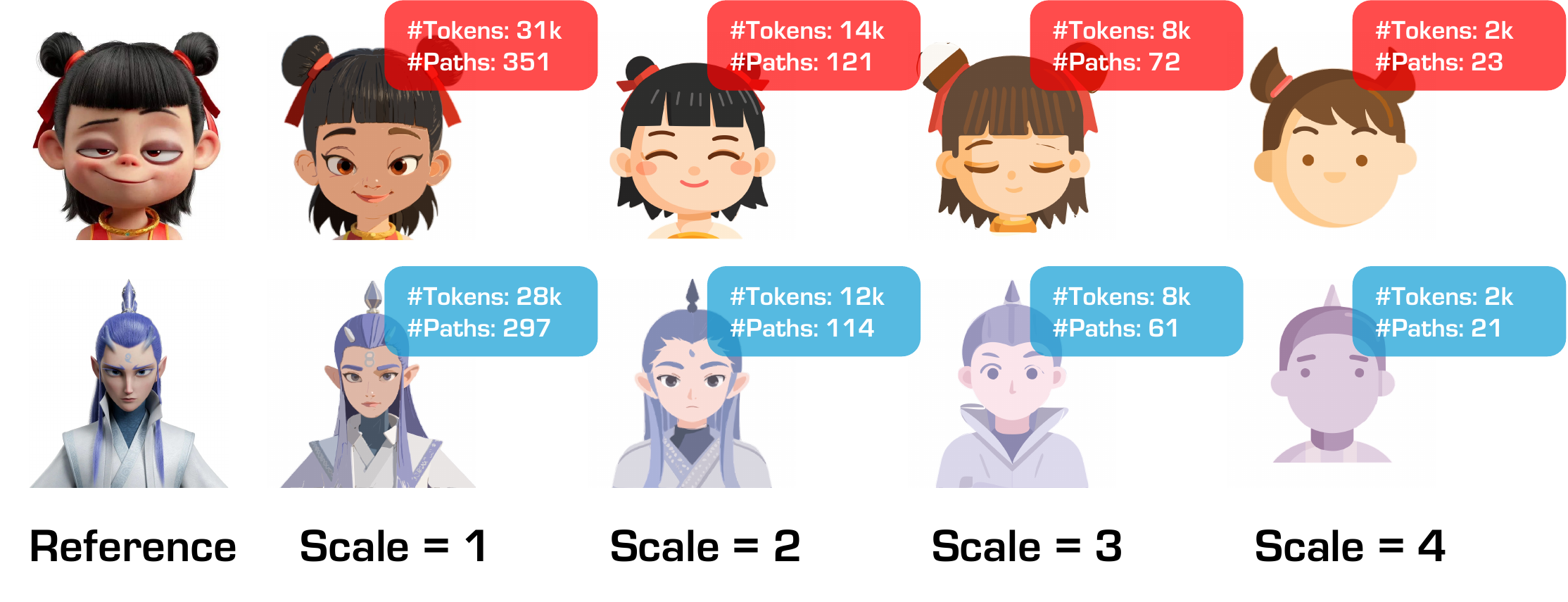}
    \caption{\small \textbf{Image Prompting Dataset Creation of MMSVG-2M Character}. By utilizing FLUX-Redux and SVG vectorization tools, image prompting data pairs can be generated. We adpot FLUX-Redux downsampling scale with $2, 3$ in practice by trading-off the character similarity and complexity of generated SVG.}
    \label{fig:redux}
\end{figure}

%% file: Figures/S1.dataset_wordcloud.tex
\begin{wrapfigure}{r}{0.5\textwidth}
    \centering
    \vspace{-5mm}
    \includegraphics[width=\linewidth]{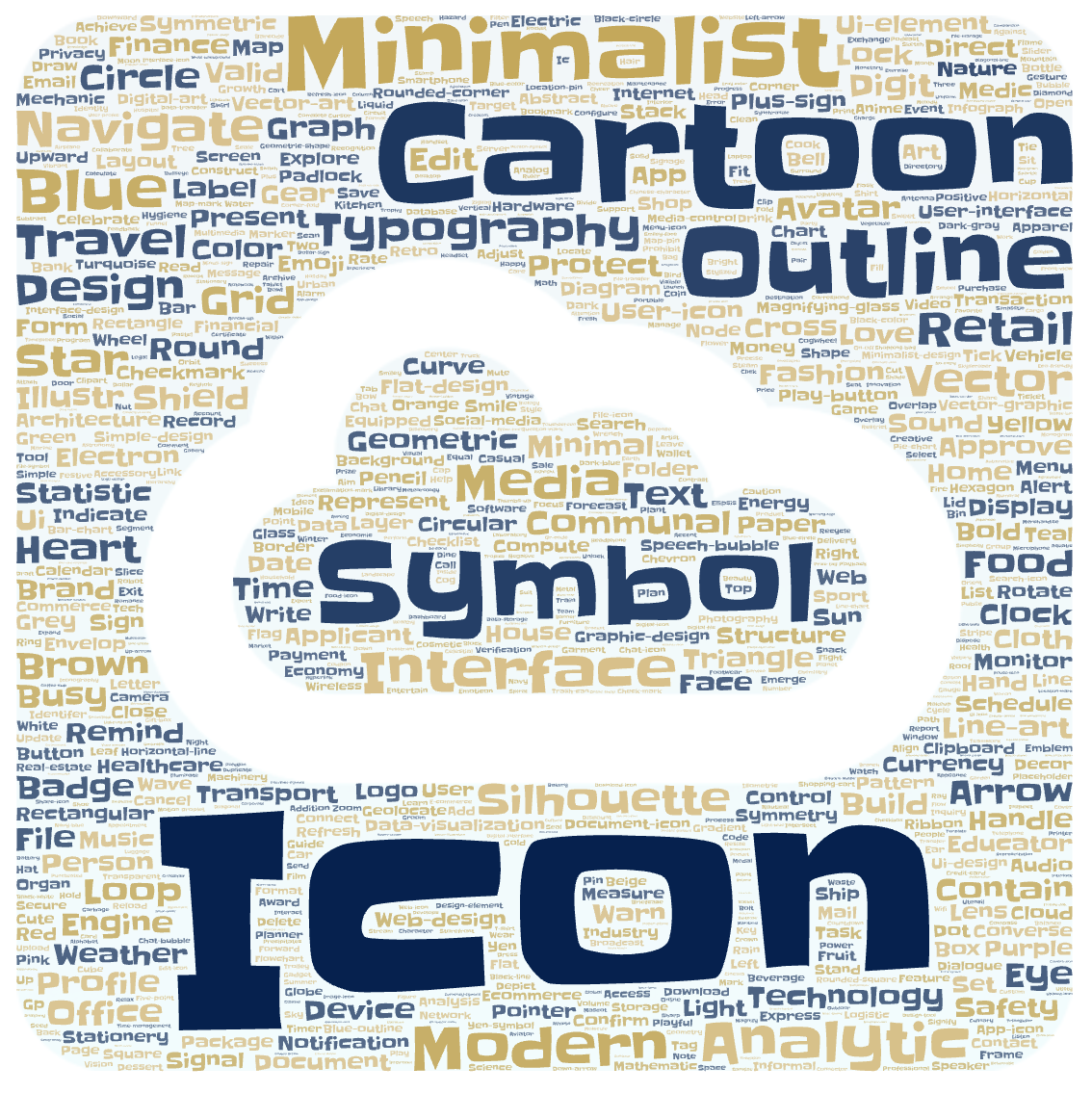}
    \caption{\small \textbf{Word Cloud Visualization of Label Distribution in the MMSVG-2M Dataset.} The size of each label corresponds to its frequency of occurrence. The larger the label, the more frequently it appears in the dataset.}
    \label{fig:wordcloud}
    \vspace{-10mm}
\end{wrapfigure}

%% file: Figures/3.scale_up.tex
\begin{figure} 
    \centering 
    \begin{subfigure}{0.49\linewidth} 
        \centering 
        \includegraphics[width=\linewidth]{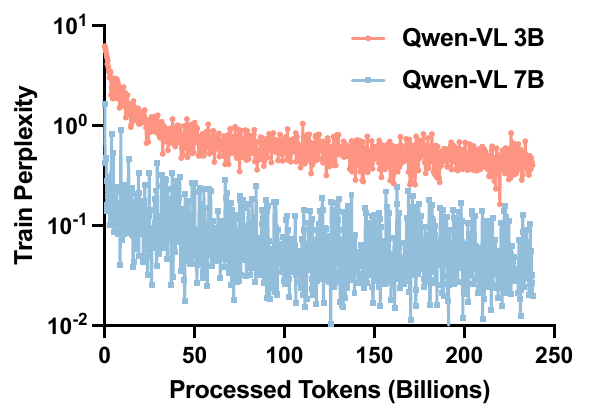} 
        \caption{Training PPL for our models.} 
    \end{subfigure} 
    \hfill 
    \begin{subfigure}{0.49\linewidth} 
        \centering 
        \includegraphics[width=\linewidth]{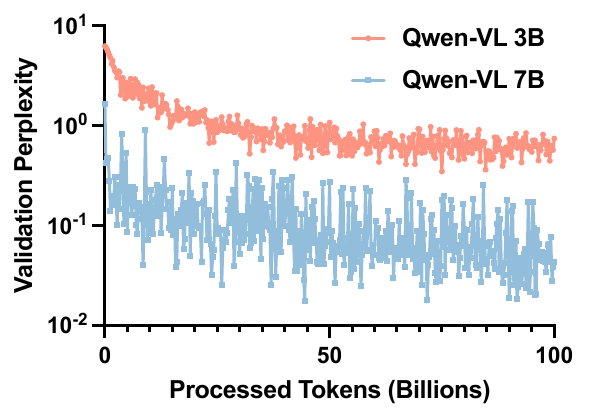} 
        \caption{Validation PPL for our models.} 
    \end{subfigure} 
    \caption{\small \textbf{Training and Validation Perplexity (PPL) for OmniSVG Models.} We train all the models from scratch on 250 billion tokens. We observe that the performance grows with model sizes.} 
    \label{fig:scale_up} 
\end{figure}

%% file: Figures/gallery.tex
\begin{figure*}
    \centering
    \includegraphics[width=0.92\linewidth]{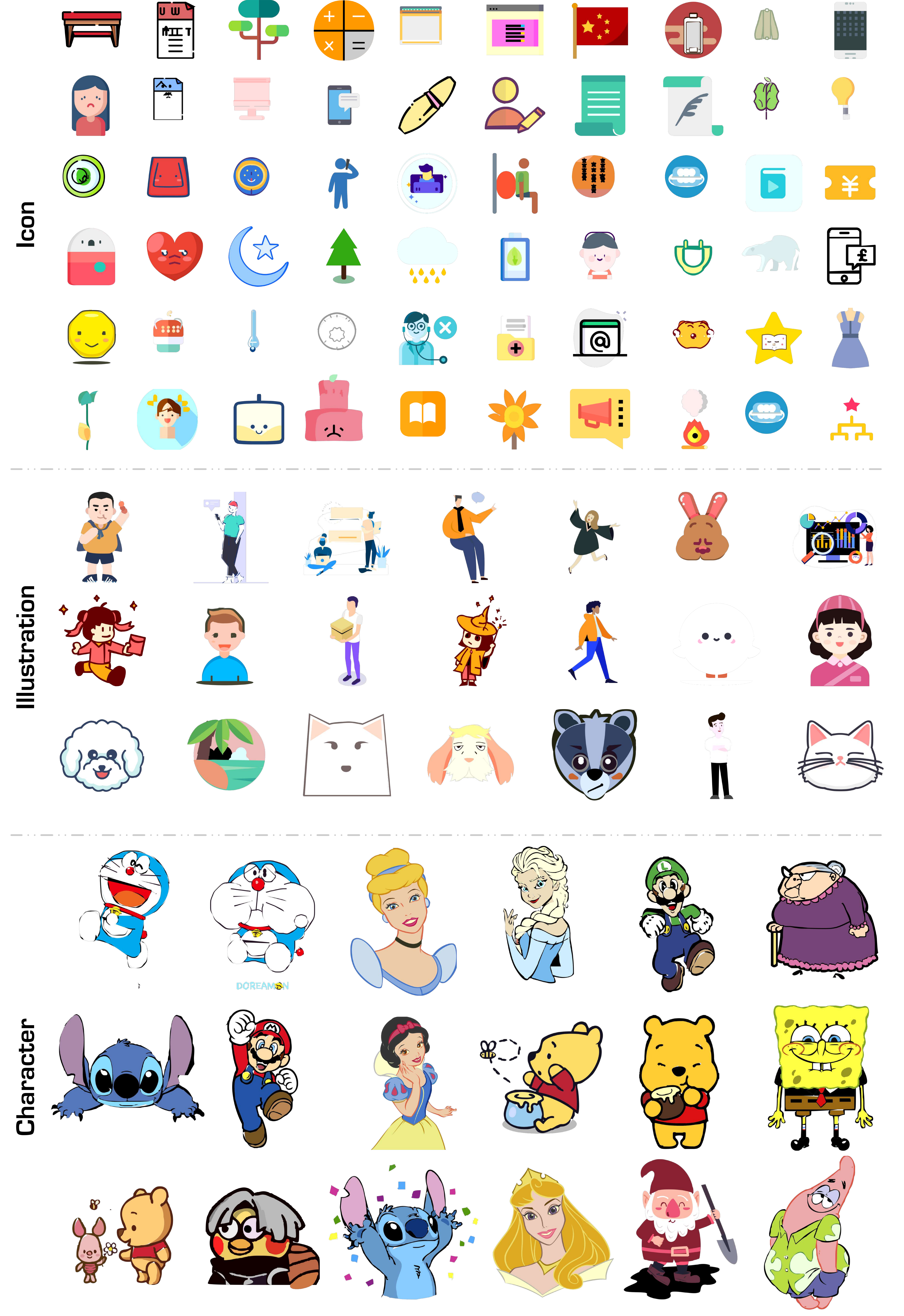}
    
    \caption{\small \textbf{Illustration of the SVG Generation Capabilities of OmniSVG.} 
    }
    \label{fig:Gallery}
\end{figure*}